\documentclass{article}

\usepackage{microtype}
\usepackage{tikz, pgfplots,graphicx,xcolor}
\usetikzlibrary{plotmarks,spy,backgrounds,positioning}
\pgfplotsset{compat=newest}
\usepackage{subfigure}
\usepackage{booktabs} % for professional tables
\usepackage[utf8]{inputenc}
\usepackage{amsmath,amssymb,amsthm}
\usepackage{siunitx}

\usepackage{hyperref}

\usepackage[accepted]{icml2018}

\newcommand{\T}{{\sf T}}

\newcommand{\I}{\mathbf{I}}
\newcommand{\V}{\mathbf{V}}
\newcommand{\X}{\mathbf{X}}
\newcommand{\Z}{\mathbf{Z}}
\newcommand{\C}{\mathbf{C}}
\newcommand{\Q}{\mathbf{Q}}
\newcommand{\M}{\mathbf{M}}

\newcommand{\e}{\mathbf{e}}
\newcommand{\w}{\mathbf{w}}
\newcommand{\x}{\mathbf{x}}
\newcommand{\y}{\mathbf{y}}
\newcommand{\z}{\mathbf{z}}
\renewcommand{\j}{\mathbf{j}}

\newcommand{\bmu}{\boldsymbol{\mu}}
\newcommand{\bLambda}{\boldsymbol{\Lambda}}

\DeclareMathOperator{\tr}{tr}

\newtheorem{Assumption}{Assumption}
\newtheorem{Theorem}{Theorem}

\newtheorem{Remark}{Remark}

\begin{document}

\twocolumn[
%\icmltitle{The Dynamics of Learning from High Dimensional Data:\\ A Random Matrix Framework}
\icmltitle{The Dynamics of Learning: A Random Matrix Approach}



\icmlsetsymbol{equal}{*}

\begin{icmlauthorlist}
\icmlauthor{Zhenyu Liao}{cs}
\icmlauthor{Romain Couillet}{cs,gstats}
\end{icmlauthorlist}

\icmlaffiliation{cs}{Laboratoire des Signaux et Systèmes (L2S), CentraleSupélec, Université Paris-Saclay, France;}

\icmlaffiliation{gstats}{G-STATS Data Science Chair, GIPSA-lab, University Grenobles-Alpes, France}

\icmlcorrespondingauthor{Zhenyu Liao}{zhenyu.liao@l2s.centralesupelec.fr}
\icmlcorrespondingauthor{Romain Couillet}{romain.couillet@centralesupelec.fr}

\icmlkeywords{Neural Networks, Gradient Descent, Random Matrix Theory, High dimensional Statistics}

\vskip 0.3in
]



\printAffiliationsAndNotice{}  % leave blank if no need to mention equal contribution

\begin{abstract}
Understanding the learning dynamics of neural networks is one of the key issues for the improvement of optimization algorithms as well as for the theoretical comprehension of why deep neural nets work so well today. In this paper, we introduce a random matrix-based framework to analyze the learning dynamics of a single-layer linear network on a binary classification problem, for data of simultaneously large dimension and size, trained by gradient descent. Our results provide rich insights into common questions in neural nets, such as overfitting, early stopping and the initialization of training, thereby opening the door for future studies of more elaborate structures and models appearing in today's neural networks.
\end{abstract}

\section{Introduction}
\label{sec:introduction}
Deep neural networks trained with backpropagation have commonly attained superhuman performance in applications of computer vision \cite{krizhevsky2012imagenet} and many others \cite{schmidhuber2015deep} and are thus receiving an unprecedented research interest. Despite the rapid growth of the list of successful applications with these gradient-based methods, our theoretical understanding, however, is progressing at a more modest pace.

One of the salient features of deep networks today is that they often have far more model parameters than the number of training samples that they are trained on, but meanwhile some of the models still exhibit remarkably good generalization performance when applied to unseen data of similar nature, while others generalize poorly in exactly the same setting. A satisfying explanation of this phenomenon would be the key to more powerful and reliable network structures.

To answer such a question, statistical learning theory has proposed interpretations from the viewpoint of system complexity \cite{vapnik2013nature,bartlett2002rademacher,poggio2004general}. In the case of large numbers of parameters, it is suggested to apply some form of regularization to ensure good generalization performance. Regularizations can be explicit, such as the dropout technique \cite{srivastava2014dropout} or the $l_2$-penalization (weight decay) as reported in \cite{krizhevsky2012imagenet}; or implicit, as in the case of the early stopping strategy \cite{yao2007early} or the stochastic gradient descent algorithm itself \cite{zhang2016understanding}.

Inspired by the recent line of works \cite{saxe2013exact,advani2017high}, in this article we introduce a random matrix framework to analyze the training and, more importantly, the generalization performance of neural networks, trained by gradient descent. Preliminary results established from a toy model of two-class classification on a single-layer linear network are presented, which, despite their simplicity, shed new light on the understanding of many important aspects in training neural nets. In particular, we demonstrate how early stopping can naturally protect the network against overfitting, which becomes more severe as the number of training sample approaches the dimension of the data. We also provide a strict lower bound on the training sample size for a given classification task in this simple setting. A byproduct of our analysis implies that random initialization, although commonly used in practice in training deep networks \cite{glorot2010understanding,krizhevsky2012imagenet}, may lead to a degradation of the network performance. 

From a more theoretical point of view, our analyses allow one to evaluate any functional of the eigenvalues of the sample covariance matrix of the data (or of the data representation learned from previous layers in a deep model), which is at the core of understanding many experimental observations in today's deep networks \cite{glorot2010understanding,ioffe2015batch}. Our results are envisioned to generalize to more elaborate settings, notably to deeper models that are trained with the stochastic gradient descent algorithm, which is of more practical interest today due to the tremendous size of the data.

\emph{Notations}: Boldface lowercase (uppercase) characters stand for vectors (matrices), and non-boldface for scalars respectively. $\mathbf{0}_p$ is the column vector of zeros of size $p$, and $\mathbf{I}_p$ the $p \times p$ identity matrix. The notation $(\cdot)^\T$ denotes the transpose operator. The norm $\| \cdot \| $ is the Euclidean norm for vectors and the operator norm for matrices. $\Im(\cdot)$ denotes the imaginary part of a complex number. For $x \in \mathbb{R}$, we denote for simplicity $(x)^+ \equiv \max(x,0)$.

In the remainder of the article, we introduce the problem of interest and recall the results of \cite{saxe2013exact} in Section~\ref{sec:problem}. After a brief overview of basic concepts and methods to be used throughout the article in Section~\ref{sec:preliminaries}, our main results on the training and generalization performance of the network are presented in Section~\ref{sec:performance}, followed by a thorough discussion in Section~\ref{sec:discuss} and experiments on the popular MNIST database \cite{lecun1998mnist} in Section~\ref{sec:validations}. Section~\ref{sec:conclusion} concludes the article by summarizing the main results and outlining future research directions.

\section{Problem Statement} 
\label{sec:problem}

Let the training data $\x_1, \ldots, \x_n \in \mathbb{R}^p$ be independent vectors drawn from two distribution classes $\mathcal{C}_1$ and $\mathcal{C}_2$ of cardinality $n_1$ and $n_2$ (thus $n_1 + n_2 = n$), respectively. We assume that the data vector $\x_i$ of class $\mathcal{C}_a$ can be written as
\[
	\x_i = (-1)^a \bmu + \z_i
\]
% \[
% 	\begin{cases}
% 		\x_i = -\bmu + \z_i, &a=1\\
% 		\x_i = \bmu + \z_i, &a=2
% 	\end{cases}
% \]
for $a = \{1,2\}$, with $\bmu \in \mathbb{R}^p$ and $\z_i$ a Gaussian random vector $\z_i \sim \mathcal{N}(\mathbf{0}_p, \I_p)$. In the context of a binary classification problem, one takes the label $y_i = -1$ for $\x_i \in \mathcal{C}_1$ and $y_j = 1$ for $\x_j \in \mathcal{C}_2$ to distinguish the two classes.

We denote the training data matrix $\X = \begin{bmatrix} \x_1, \ldots, \x_n \end{bmatrix} \in \mathbb{R}^{p \times n}$ by cascading all $\x_i$'s as column vectors and associated label vector $\y \in \mathbb{R}^n$. With the pair $\{\X, \y\}$, a classifier is trained using ``full-batch'' gradient descent to minimize the loss function $L(\w)$ given by
\[
	L(\w) = \frac1{2n} \| \y^\T - \w^\T \X \|^2
\]
so that for a new datum $\hat \x$, the output of the classifier is $\hat y = \w^\T \hat \x$, the sign of which is then used to decide the class of $\hat \x$. The derivative of $L$ with respective to $\w$ is given by
\[
	\frac{\partial L(\w)}{\partial \w} = - \frac1{n} \X (\y - \X^\T \w).
\]

The gradient descent algorithm \cite{boyd2004convex} takes small steps of size $\alpha$ along the \emph{opposite direction} of the associated gradient, i.e., $\w_{t+1} = \w_t - \alpha \frac{\partial L(\w)}{\partial \w} \big|_{\w = \w_t}$.

Following the previous works of \cite{saxe2013exact,advani2017high}, when the learning rate $\alpha$ is small, $\w_{t+1}$ and $\w_t$ are close to each other so that by performing a continuous-time approximation, one obtains the following differential equation
\[
	\frac{\partial \w(t)}{\partial t} = - \alpha \frac{\partial L(\w)}{\partial \w} = \frac{\alpha}{n} \X \left(\y - \X^\T \w(t) \right)
\]
the solution of which is given explicitly by
% \[
% 	\frac{\partial \left( ( \X \X^\T )^{-1} \X \y - \w(t) \right)}{\partial t} = - \frac{\alpha}n \X \X^\T \left( ( \X \X^\T )^{-1} \X \y - \w(t) \right)
% \]
% \[
% 	( \X \X^\T )^{-1} \X \y - \w(t) = \exp \left( - \frac{\alpha t}n \X \X^\T \right) \left( ( \X \X^\T )^{-1} \X \y - \w_0 \right)
% \]
% \[
% 	( \X \X^\T )^{-1} \X \y - \w(t) =  e^{ - \frac{\alpha t}n \X \X^\T} \left( ( \X \X^\T )^{-1} \X \y - \w_0 \right)
% \]
\begin{equation}
	\w(t) = e^{- \frac{\alpha t}n \X \X^\T } \w_0 + \left(\I_p - e^{- \frac{\alpha t}n \X\X^\T } \right) ( \X\X^\T )^{-1} \X\y
	\label{eq:solution-de}
\end{equation}
if one assumes that $\X \X^\T$ is invertible (only possible in the case $p < n$), with $\w_0 \equiv \w(t=0)$ the initialization of the weight vector; we recall the definition of the exponential of a matrix $\frac1n \X \X^\T$ given by the power series $e^{\frac1n \X \X^\T} = \sum_{k=0}^\infty \frac1{k!} (\frac1n \X \X^\T)^k = \V e^{\bLambda} \V^\T$, with the eigendecomposition of $\frac1n \X \X^\T = \V \bLambda \V^\T$ and $e^{\bLambda}$ is a diagonal matrix with elements equal to the exponential of the elements of $\bLambda$. As $t\to \infty$ the network ``forgets'' the initialization $\w_0$ and results in the least-square solution $\w_{LS} \equiv ( \X\X^\T )^{-1} \X\y$.

When $p>n$, $\X\X^\T$ is no longer invertible. Assuming $\X^\T \X$ is invertible and writing $\X\y = \left(\X\X^\T\right)\X\left(\X^\T\X\right)^{-1}\y$, the solution is similarly given by
\[
	\w(t) = e^{- \frac{\alpha t}n \X \X^\T } \w_0 + \X \left(\I_n - e^{- \frac{\alpha t}n \X^\T \X } \right) ( \X^\T \X )^{-1} \y
\]
with the least-square solution $\w_{LS} \equiv \X( \X^\T\X )^{-1} \y$. 

In the work of \cite{advani2017high} it is assumed that $\X$ has i.i.d.\@ entries and that there is no linking structure between the data and associated targets in such a way that the ``true'' weight vector $\bar \w$ to be learned is independent of $\X$ so as to simplify the analysis. In the present work we aim instead at exploring the capacity of the network to retrieve the (mixture modeled) data structure and position ourselves in a more realistic setting where $\w$ captures the different statistical structures (between classes) of the pair $(\X,\y)$. Our results are thus of more guiding significance for practical interests.

From \eqref{eq:solution-de} note that both $e^{-\frac{\alpha t}{n} \X \X^\T}$ and $\I_p - e^{-\frac{\alpha t}{n} \X \X^\T}$ share the same eigenvectors with the \emph{sample covariance matrix} $\frac1n \X \X^\T$, which thus plays a pivotal role in the network learning dynamics. More concretely, the projections of $\w_0$ and $\w_{LS}$ onto the eigenspace of $\frac1n \X \X^\T$, weighted by functions ($\exp(-\alpha t \lambda_i)$ or $1-\exp(-\alpha t \lambda_i)$) of the associated eigenvalue $\lambda_i$, give the temporal evolution of $\w(t)$ and consequently the training and generalization performance of the network. The core of our study therefore consists in deeply understanding of the eigenpairs of this sample covariance matrix, which has been largely investigated in the random matrix literature \cite{bai2010spectral}.

\section{Preliminaries}
\label{sec:preliminaries}
Throughout this paper, we will be relying on some basic yet powerful concepts and methods from random matrix theory, which shall be briefly highlighted in this section.

\subsection{Resolvent and deterministic equivalents}
\label{subsec:resolvent-and-its-D-E}

Consider an $n \times n$ Hermitian random matrix $\M$. We define its \emph{resolvent} $\Q_{\M}(z)$, for $z \in \mathbb{C}$ not an eigenvalue of $\M$, as
\[
	\Q_{\M}(z) = \left( \M - z \mathbf{I}_n \right)^{-1}.
\]
Through the Cauchy integral formula discussed in the following subsection, as well as its central importance in random matrix theory, $\Q_{\frac1n \X\X^\T}(z)$ is the key object investigated in this article.

For certain simple distributions of $\M$, one may define a so-called \emph{deterministic equivalent} \cite{hachem2007deterministic,couillet2011random} $\bar\Q_{\M}$ for $\Q_{\M}$, which is a deterministic matrix such that for all $\mathbf{A}\in \mathbb{R}^{n \times n}$ and all $\mathbf{a,b} \in \mathbb{R}^n$ of bounded (spectral and Euclidean, respectively) norms, $\frac1n \tr \left( \mathbf{A} \Q_{\M} \right) - \frac1n \tr \left( \mathbf{A}  \bar \Q_{\M} \right) \to 0$ and $\mathbf{a}^\T \left( \Q_{\M} - \bar \Q_{\M} \right) \mathbf{b} \to 0$ almost surely as $n \to \infty$. As such, deterministic equivalents allow to transfer random spectral properties of $\M$ in the form of deterministic limiting quantities and thus allows for a more detailed investigation.

\subsection{Cauchy's integral formula}
\label{subsec:cauchy-integral-and-residue}

First note that the resolvent $\Q_{\M}(z)$ has the same eigenspace as $\M$, with associated eigenvalue $\lambda_i$ replaced by $\frac1{\lambda_i - z}$. As discussed at the end of Section~\ref{sec:problem}, our objective is to evaluate functions of these eigenvalues, which reminds us of the fundamental Cauchy's integral formula, stating that for any function $f$ holomorphic on an open subset $U$ of the complex plane, one can compute $f(\lambda)$ by contour integration. More concretely, for a closed positively (counter-clockwise) oriented path $\gamma$ in $U$ with winding number one (i.e., describing a $360^\circ$ rotation), one has, for $\lambda$ contained in the surface described by $\gamma$, $\frac1{2\pi i} \oint_{\gamma} \frac{f(z)}{z - \lambda} dz = f(\lambda)$ and $\frac1{2\pi i} \oint_{\gamma} \frac{f(z)}{z - \lambda} dz = 0$ if $\lambda$ lies outside the contour of $\gamma$.

With Cauchy's integral formula, one is able to evaluate more sophisticated functionals of the random matrix $\M$. For example, for $f(\M) \equiv \mathbf{a}^\T e^{\M} \mathbf{b}$ one has
\[
	f(\M) = -\frac1{2 \pi i} \oint_{\gamma} \exp(z) \mathbf{a}^\T \Q_{\M}(z) \mathbf{b}\ dz
\]
with $\gamma$ a positively oriented path circling around \emph{all} the eigenvalues of $\M$. Moreover, from the previous subsection one knows that the bilinear form $\mathbf{a}^\T \Q_{\M}(z) \mathbf{b}$ is asymptotically close to a non-random quantity $\mathbf{a}^\T \bar \Q_{\M}(z) \mathbf{b}$. One thus deduces that the functional $\mathbf{a}^\T e^{\M} \mathbf{b}$ has an asymptotically deterministic behavior that can be expressed as $-\frac1{2 \pi i} \oint_{\gamma} \exp(z) \mathbf{a}^\T \bar \Q_{\M}(z) \mathbf{b}\ dz$. 

This observation serves in the present article as the foundation for the performance analysis of the gradient-based classifier, as described in the following section.

\section{Temporal Evolution of Training and Generalization Performance}
\label{sec:performance}

With the explicit expression of $\w(t)$ in \eqref{eq:solution-de}, we now turn our attention to the training and generalization performances of the classifier as a function of the training time $t$. To this end, we shall be working under the following assumptions.
\begin{Assumption}[Growth Rate]
As $n \to \infty$,
\begin{enumerate}
	\item $\frac{p}{n} \to c  \in (0,\infty)$.
	\item For $a= \{1,2\}$, $\frac{n_a}{n} \to c_a \in (0,1)$.
	\item $\| \bmu \| = O(1)$.
	% \item $\| \bmu_a \| = O(1)$
	% \item $\| \C_a \| = O(1)$.
\end{enumerate}
\label{ass:growth-rate}
\end{Assumption}
The above assumption ensures that the matrix $ \frac1n \X \X^\T$ is of bounded operator norm for all large $n,p$ with probability one \cite{bai1998no}. 

\begin{Assumption}[Random Initialization]
Let $\w_0 \equiv \w(t=0)$ be a random vector with i.i.d.\@ entries of zero mean, variance $\sigma^2/p$ for some $\sigma>0$ and finite fourth moment.
\label{ass:initialization}
\end{Assumption}

We first focus on the generalization performance, i.e., the average performance of the trained classifier taking as input an unseen new datum $\hat \x$ drawn from class $\mathcal{C}_1$ or $\mathcal{C}_2$.

\subsection{Generalization Performance} 
\label{subsec:generalization-perf}

To evaluate the generalization performance of the classifier, we are interested in two types of misclassification rates, for a new datum $\hat \x$ drawn from class $\mathcal{C}_1$ or $\mathcal{C}_2$, as
\[
	{\rm P}( \w(t)^\T \hat \x > 0~|~\hat \x \in \mathcal{C}_1), \quad {\rm P}( \w(t)^\T \hat \x < 0~|~\hat \x \in \mathcal{C}_2).
\]

Since the new datum $\hat \x$ is independent of $\w(t)$, $\w(t)^\T \hat \x$ is a Gaussian random variable of mean $\pm \w(t)^\T \bmu$ and variance $ \| \w(t) \|^2 $. The above probabilities can therefore be given via the $Q$-function: $Q(x) \equiv \frac1{\sqrt{2\pi}} \int_x^{\infty} \exp\left( -\frac{u^2}2 \right) du$. We thus resort to the computation of $\w(t)^\T \bmu$ as well as $ \w(t)^\T \w(t) $ to evaluate the aforementioned classification error.

For $\bmu^\T \w(t)$, with Cauchy's integral formula we have
\begin{align*}
	&\bmu^\T \w(t) = \bmu^\T e^{- \frac{\alpha t}n \X \X^\T } \w_0 + \bmu^\T \left(\I_p - e^{- \frac{\alpha t}n \X\X^\T } \right) \w_{LS}\\
	&= -\frac1{2\pi i} \oint_{\gamma} f_t(z) \bmu^\T \left( \frac1n \X \X^\T - z \I_p \right)^{-1} \w_0 \ dz \\ 
	& -\frac1{2\pi i} \oint_{\gamma} \frac{1-f_t(z)}{z} \bmu^\T \left( \frac1n \X \X^\T - z \I_p \right)^{-1} \frac1n \X\y \ dz
\end{align*}
with $f_t(z) \equiv \exp(-\alpha t z)$, for a positive closed path $\gamma$ circling around all eigenvalues of $\frac1n \X \X^\T$. Note that the data matrix $\X$ can be rewritten as
\[
	\X = -\bmu \j_1^\T + \bmu \j_2^\T + \Z = \bmu \y^\T + \Z
\]
with $\Z \equiv \begin{bmatrix} \z_1, \ldots, \z_n \end{bmatrix} \in \mathbb{R}^{p \times n}$ of i.i.d.\@ $\mathcal{N}(0,1)$ entries and $\j_a \in \mathbb{R}^n$ the canonical vectors of class $\mathcal{C}_a$ such that $(\j_a)_i = \delta_{\x_i \in \mathcal{C}_a}$. To isolate the deterministic vectors $\bmu$ and $\j_a$'s from the random $\Z$ in the expression of $\bmu^\T \w(t)$, we exploit Woodbury's identity to obtain
\begin{align*}
	&\left( \frac1n \X \X^\T - z \I_p \right)^{-1} = \Q(z) - \Q(z) \begin{bmatrix} \bmu & \frac1n \Z\y \end{bmatrix} \\ 
	&\begin{bmatrix} \bmu^\T \Q(z) \bmu & 1+\frac1n \bmu^\T \Q(z) \Z\y \\ * & -1 + \frac1n \y^\T \Z^\T \Q(z) \frac1n \Z \y \end{bmatrix}^{-1} \begin{bmatrix} \bmu^\T \\ \frac1n \y^\T \Z^\T \end{bmatrix} \Q(z)
\end{align*}
where we denote the resolvent $\Q(z) \equiv \left( \frac1n \Z\Z^\T - z \I_p \right)^{-1}$, a deterministic equivalent of which is given by
\[
	\Q(z) \leftrightarrow \bar \Q(z) \equiv m(z) \I_p
\]
with $m(z)$ determined by the popular Marčenko–Pastur equation \cite{marvcenko1967distribution}
\begin{equation}
	%m(z) = \frac1{1 - c - z - czm(z)}
	m(z) = \frac{1-c-z}{2cz} + \frac{\sqrt{(1-c-z)^2 - 4cz}}{2cz}
	\label{eq:MP-equation}
\end{equation}
%where the sign in front of $\frac{\sqrt{(1-c-z)^2 - 4cz}}{2cz}$ is chosen such that $\Im(z) \cdot \Im m(z) >0$.
where the branch of the square root is selected in such a way that $\Im(z) \cdot \Im m(z) >0$, i.e., for a given $z$ there exists a \emph{unique} corresponding $m(z)$.

Substituting $\Q(z)$ by the simple form deterministic equivalent $m(z) \I_p$, we are able to estimate the random variable $\bmu^\T \w(t)$ with a contour integral of some deterministic quantities as $n,p \to \infty$. Similar arguments also hold for $\w(t)^\T \w(t)$, together leading to the following theorem.

\begin{Theorem}[Generalization Performance]
Let Assumptions~\ref{ass:growth-rate} and~\ref{ass:initialization} hold. As $n \to \infty$, with probability one
\begin{align*}
	&{\rm P}( \w(t)^\T \hat \x > 0~|~\hat \x \in \mathcal{C}_1) - Q\left( \frac{E }{ \sqrt{V} } \right) \to 0 \\
	&{\rm P}( \w(t)^\T \hat \x < 0~|~\hat \x \in \mathcal{C}_2) - Q\left( \frac{E }{ \sqrt{V} } \right)\to 0
\end{align*}
where
\begin{align*}
	E &\equiv -\frac1{2\pi i} \oint_{\gamma} \frac{1-f_t(z)}{z} \frac{ \| \bmu \|^2 m(z) \ dz}{ \left( \| \bmu \|^2 +c \right) m(z) +1 } \\
	V &\equiv \frac1{2\pi i} \oint_{\gamma} \left[\frac{ \frac1{z^2} \left(1-f_t(z)\right)^2\ }{ \left( \| \bmu \|^2 +c \right) m(z) +1 } - \sigma^2 f_t^2(z) m(z) \right]dz
\end{align*}
with $\gamma$ a closed positively oriented path that contains all eigenvalues of $\frac1n \X \X^\T$ and the origin, $f_t(z) \equiv \exp(-\alpha t z)$ and $m(z)$ given by Equation~\eqref{eq:MP-equation}.
\label{theo:generalize-perf}
\end{Theorem}

Although derived from the case $p<n$, Theorem~\ref{theo:generalize-perf} also applies when $p>n$. To see this, note that with Cauchy's integral formula, for $z\neq0$ not an eigenvalue of $\frac1n \X \X^\T$ (thus not of  $\frac1n \X^\T \X$), one has $\X \left( \frac1n \X^\T \X - z \I_n \right)^{-1}\y = \left( \frac1n \X\X^\T - z \I_p \right)^{-1} \X \y$, which further leads to the same expressions as in Theorem~\ref{theo:generalize-perf}. Since $\frac1n \X\X^\T$ and $\frac1n \X^\T\X$ have the same eigenvalues except for additional zero eigenvalues for the larger matrix, the path $\gamma$ remains unchanged (as we demand that $\gamma$ contains the origin) and hence Theorem~\ref{theo:generalize-perf} holds true for both $p<n$ and $p>n$. The case $p=n$ can be obtained by continuity arguments.

\subsection{Training performance}
\label{eq:training-perf}

To compare generalization versus training performance, we are now interested in the behavior of the classifier when applied to the training set $\X$. To this end, we consider the random vector $\X^\T \w(t)$ given by
\[
	\X^\T \w(t) = \X^\T e^{- \frac{\alpha t}n \X \X^\T } \w_0 + \X^\T \left(\I_p - e^{- \frac{\alpha t}n \X\X^\T } \right) \w_{LS}.
\]

Note that the $i$-th entry of $\X^\T \w(t)$ is given by the bilinear form $\e_i^\T \X^\T \w(t)$, with $\e_i$ the canonical vector with unique non-zero entry $[\e_i]_i = 1$. With previous notations we have
\begin{align*}
	&\e_i^\T \X^\T \w(t)\\ 
	&= -\frac1{2\pi i} \oint_{\gamma} f_t(z, t) \e_i^\T \X^\T \left( \frac1n \X \X^\T - z \I_p \right)^{-1} \w_0 \ dz\\
	&-\frac1{2\pi i} \oint_{\gamma} \frac{1-f_t(z)}{z} \e_i^\T \frac1n \X^\T \left( \frac1n \X \X^\T - z \I_p \right)^{-1} \X \y \ dz
\end{align*}
which yields the following results.

\begin{Theorem}[Training Performance]
Under the assumptions and notations of Theorem~\ref{theo:generalize-perf}, as $n \to \infty$,
\begin{align*}
	&{\rm P}( \w(t)^\T \x_i > 0~|~\x_i \in \mathcal{C}_1) - Q\left( \frac{E_* }{ \sqrt{V_* - E_*^2 } } \right) \to 0 \\
	&{\rm P}( \w(t)^\T \x_i < 0~|~\x_i \in \mathcal{C}_2) - Q\left( \frac{E_* }{ \sqrt{V_* - E_*^2} } \right)\to 0
\end{align*}
almost surely, with
\begin{align*}
	E_* &\equiv \frac1{2\pi i} \oint_{\gamma} \frac{1-f_t(z)}{z} \frac{dz}{ \left( \| \bmu \|^2 +c \right) m(z) +1 } \\
	V_* &\equiv \frac1{2\pi i} \oint_{\gamma} \left[\frac{ \frac1{z} \left(1-f_t(z)\right)^2\ }{ \left( \| \bmu \|^2 +c \right) m(z) +1 } - \sigma^2 f_t^2(z) z m(z) \right] dz.
\end{align*}
\label{theo:training-perf}
\end{Theorem}

\begin{figure}[htb]
\vskip 0.1in
\begin{center}
\begin{tikzpicture}[font=\footnotesize,spy using outlines]
\renewcommand{\axisdefaulttryminticks}{4} 
\pgfplotsset{every major grid/.style={densely dashed}}       
\tikzstyle{every axis y label}+=[yshift=-10pt] 
\tikzstyle{every axis x label}+=[yshift=5pt]
\pgfplotsset{every axis legend/.append style={cells={anchor=west},fill=white, at={(0.98,0.98)}, anchor=north east, font=\footnotesize, }}
\begin{axis}[
width=\columnwidth,
height=0.7\columnwidth,
xmin=0,
xmax=300,
ymin=-.01,
ymax=.5,
xlabel={Training time $(t)$},
ylabel={Misclassification rate},
ytick={0,0.1,0.2,0.3,0.4,0.5},
grid=major,
%ymajorgrids=false,
scaled ticks=true,
]
\addplot[color=blue!60!white,line width=1pt] coordinates{
(0.000000,0.482031)(6.000000,0.211094)(12.000000,0.107344)(18.000000,0.064687)(24.000000,0.046250)(30.000000,0.031094)(36.000000,0.023125)(42.000000,0.017500)(48.000000,0.014375)(54.000000,0.012344)(60.000000,0.010313)(66.000000,0.008906)(72.000000,0.007656)(78.000000,0.006562)(84.000000,0.005938)(90.000000,0.005469)(96.000000,0.005000)(102.000000,0.004687)(108.000000,0.004531)(114.000000,0.004063)(120.000000,0.003750)(126.000000,0.003594)(132.000000,0.003125)(138.000000,0.002969)(144.000000,0.002656)(150.000000,0.002656)(156.000000,0.002344)(162.000000,0.002344)(168.000000,0.002188)(174.000000,0.002031)(180.000000,0.002031)(186.000000,0.002031)(192.000000,0.001875)(198.000000,0.001875)(204.000000,0.001875)(210.000000,0.001875)(216.000000,0.001719)(222.000000,0.001719)(228.000000,0.001719)(234.000000,0.001719)(240.000000,0.001563)(246.000000,0.001563)(252.000000,0.001563)(258.000000,0.001563)(264.000000,0.001563)(270.000000,0.001563)(276.000000,0.001563)(282.000000,0.001563)(288.000000,0.001563)(294.000000,0.001563)
};
\addlegendentry{{ Simulation: training performance }};
\addplot+[only marks,mark = x,color=blue!60!white] coordinates{
(0.000000,0.500000)(6.000000,0.228091)(12.000000,0.109439)(18.000000,0.063533)(24.000000,0.042920)(30.000000,0.031943)(36.000000,0.025256)(42.000000,0.020767)(48.000000,0.017539)(54.000000,0.015102)(60.000000,0.013199)(66.000000,0.011674)(72.000000,0.010429)(78.000000,0.009398)(84.000000,0.008534)(90.000000,0.007802)(96.000000,0.007177)(102.000000,0.006639)(108.000000,0.006172)(114.000000,0.005765)(120.000000,0.005407)(126.000000,0.005091)(132.000000,0.004809)(138.000000,0.004558)(144.000000,0.004333)(150.000000,0.004129)(156.000000,0.003945)(162.000000,0.003777)(168.000000,0.003623)(174.000000,0.003482)(180.000000,0.003352)(186.000000,0.003232)(192.000000,0.003121)(198.000000,0.003017)(204.000000,0.002920)(210.000000,0.002829)(216.000000,0.002744)(222.000000,0.002664)(228.000000,0.002589)(234.000000,0.002519)(240.000000,0.002452)(246.000000,0.002390)(252.000000,0.002331)(258.000000,0.002276)(264.000000,0.002224)(270.000000,0.002176)(276.000000,0.002131)(282.000000,0.002090)(288.000000,0.002052)(294.000000,0.002018)
};
\addlegendentry{{ Theory: training performance }};
\addplot[densely dashed,color=red!60!white,line width=1pt] coordinates{
(0.000000,0.491875)(6.000000,0.258594)(12.000000,0.146250)(18.000000,0.101250)(24.000000,0.078594)(30.000000,0.069062)(36.000000,0.060625)(42.000000,0.056875)(48.000000,0.053594)(54.000000,0.052969)(60.000000,0.051875)(66.000000,0.050937)(72.000000,0.049688)(78.000000,0.049375)(84.000000,0.048906)(90.000000,0.048594)(96.000000,0.047813)(102.000000,0.047500)(108.000000,0.047813)(114.000000,0.048281)(120.000000,0.047813)(126.000000,0.048281)(132.000000,0.048750)(138.000000,0.048594)(144.000000,0.049063)(150.000000,0.049219)(156.000000,0.049844)(162.000000,0.049531)(168.000000,0.049063)(174.000000,0.050156)(180.000000,0.050313)(186.000000,0.050781)(192.000000,0.051406)(198.000000,0.051719)(204.000000,0.052031)(210.000000,0.052187)(216.000000,0.051875)(222.000000,0.052031)(228.000000,0.052344)(234.000000,0.052500)(240.000000,0.052344)(246.000000,0.052500)(252.000000,0.052656)(258.000000,0.052656)(264.000000,0.053281)(270.000000,0.053906)(276.000000,0.053750)(282.000000,0.054375)(288.000000,0.054531)(294.000000,0.054844)
};
\addlegendentry{{ Simulation: generalization performance }};
\addplot+[only marks,mark = o,color=red!60!white] coordinates{
(0.000000,0.500000)(6.000000,0.259797)(12.000000,0.149456)(18.000000,0.103028)(24.000000,0.081019)(30.000000,0.069156)(36.000000,0.062103)(42.000000,0.057609)(48.000000,0.054610)(54.000000,0.052548)(60.000000,0.051108)(66.000000,0.050101)(72.000000,0.049404)(78.000000,0.048937)(84.000000,0.048644)(90.000000,0.048486)(96.000000,0.048433)(102.000000,0.048463)(108.000000,0.048559)(114.000000,0.048710)(120.000000,0.048905)(126.000000,0.049135)(132.000000,0.049396)(138.000000,0.049681)(144.000000,0.049988)(150.000000,0.050311)(156.000000,0.050649)(162.000000,0.051000)(168.000000,0.051362)(174.000000,0.051732)(180.000000,0.052111)(186.000000,0.052497)(192.000000,0.052889)(198.000000,0.053287)(204.000000,0.053689)(210.000000,0.054096)(216.000000,0.054507)(222.000000,0.054921)(228.000000,0.055339)(234.000000,0.055760)(240.000000,0.056184)(246.000000,0.056610)(252.000000,0.057038)(258.000000,0.057468)(264.000000,0.057900)(270.000000,0.058332)(276.000000,0.058764)(282.000000,0.059196)(288.000000,0.059627)(294.000000,0.060056)
};
\addlegendentry{{ Theory: generalization performance }};
\begin{scope}
    \spy[black!50!white,size=1.6cm,circle,connect spies,magnification=2] on (1,0.4) in node [fill=none] at (4,1.5);
\end{scope}
\end{axis}
\end{tikzpicture}
\caption{Training and generalization performance for $\bmu = [2;\mathbf{0}_{p-1}]$, $p=256$, $n=512$, $\sigma^2 =0.1$, $\alpha = 0.01$ and $c_1 = c_2 = 1/2$. Results obtained by averaging over $50$ runs.}
\label{fig:train-and-general-perf}
\end{center}
\vskip -0.1in
\end{figure}
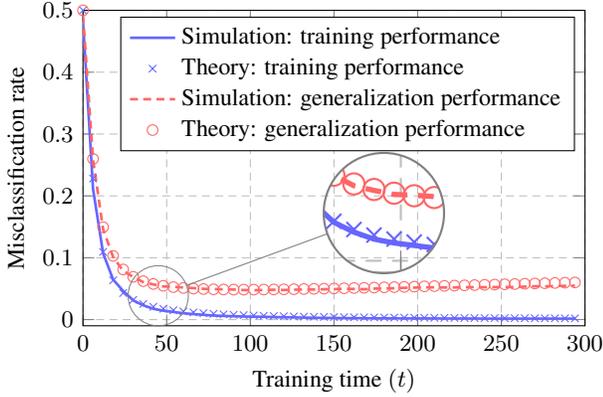

In Figure~\ref{fig:train-and-general-perf} we compare finite dimensional simulations with theoretical results obtained from Theorem~\ref{theo:generalize-perf}~and~\ref{theo:training-perf} and observe a very close match, already for not too large $n,p$. As $t$ grows large, the generalization error first drops rapidly with the training error, then goes up, although slightly, while the training error continues to decrease to zero. This is because the classifier starts to over-fit the training data $\X$ and performs badly on unseen ones. To avoid over-fitting, one effectual approach is to apply regularization strategies \cite{bishop2007pattern}, for example, to ``early stop'' (at $t=100$ for instance in the setting of Figure~\ref{fig:train-and-general-perf}) in the training process. However, this introduces new hyperparameters such as the optimal stopping time $t_{opt}$ that is of crucial importance for the network performance and is often tuned through cross-validation in practice. Theorem~\ref{theo:generalize-perf} and~\ref{theo:training-perf} tell us that the training and generalization performances, although being random themselves, have asymptotically deterministic behaviors described by $(E_*, V_*)$ and $(E, V)$, respectively, which allows for a deeper understanding on the choice of $t_{opt}$, since $E, V$ are in fact functions of $t$ via $f_t(z) \equiv \exp(-\alpha t z)$.
%or to add $L_2$ regularization to the loss function that penalizes the norm of $\w(t)$. However, this introduces new hyper-parameters such as the optimal stopping time $t_{opt}$ and the regularization parameter $\beta$, which are of crucial significance and are often tuned through cross-validation in practice. Theorem~\ref{theo:generalize-perf} and~\ref{theo:training-perf} tell us that the training and generalization performances, although being random themselves, have asymptotically deterministic behaviors on function of $(E, V)$ and $(E_*, V_*)$, respectively, which allows for a deeper understanding on the choice of $t_{opt}$, since $E, V$ are in fact functions of $t$, through $f_t(z) \equiv \exp(-\alpha t z)$.

\begin{figure}[htb]
\vskip 0.1in
\begin{center}
\begin{tikzpicture}[font=\footnotesize,spy using outlines]
\renewcommand{\axisdefaulttryminticks}{4} 
\pgfplotsset{every major grid/.style={densely dashed}}       
\tikzstyle{every axis y label}+=[yshift=-10pt] 
\tikzstyle{every axis x label}+=[yshift=5pt]
\pgfplotsset{every axis legend/.append style={cells={anchor=west},fill=white, at={(0.98,0.98)}, anchor=north east, font=\footnotesize }}
\begin{axis}[
width=\columnwidth,
height=0.7\columnwidth,
xmin=-.1,
ymin=0,
xmax=4,
ymax=.9,
yticklabels={},
bar width=4pt,
grid=major,
ymajorgrids=false,
scaled ticks=true,
]
\addplot+[ybar,mark=none,color=white,fill=blue!60!white,area legend] coordinates{
(0.000000,0.021551)(0.090628,0.797386)(0.181256,0.862039)(0.271885,0.797386)(0.362513,0.689631)(0.453141,0.646529)(0.543769,0.581876)(0.634397,0.560325)(0.725026,0.495672)(0.815654,0.431019)(0.906282,0.474121)(0.996910,0.387917)(1.087538,0.366366)(1.178166,0.366366)(1.268795,0.323264)(1.359423,0.323264)(1.450051,0.301714)(1.540679,0.301714)(1.631307,0.280163)(1.721936,0.237061)(1.812564,0.215510)(1.903192,0.215510)(1.993820,0.215510)(2.084448,0.193959)(2.175077,0.172408)(2.265705,0.172408)(2.356333,0.150857)(2.446961,0.129306)(2.537589,0.107755)(2.628218,0.086204)(2.718846,0.064653)(2.809474,0.043102)(2.900102,0.000000)(2.990730,0.000000)(3.081359,0.000000)(3.171987,0.000000)(3.262615,0.000000)(3.353243,0.000000)(3.443871,0.000000)(3.534499,0.000000)(3.625128,0.000000)(3.715756,0.000000)(3.806384,0.000000)(3.897012,0.021551)(3.987640,0.000000)(4.078269,0.000000)(4.168897,0.000000)(4.259525,0.000000)(4.350153,0.000000)(4.440781,0.000000)
};
\addlegendentry{{Eigenvalues of $\frac1n \X\X^\T$}};
\addplot[color=red!60!white,line width=1.5pt] coordinates{
(0.085786, 0.000000)(0.100000, 0.636614)(0.114213, 0.786276)(0.128426, 0.854235)(0.142639, 0.885830)(0.156852, 0.898331)(0.171066, 0.899981)(0.185279, 0.895191)(0.199492, 0.886499)(0.213705, 0.875437)(0.227918, 0.862966)(0.242132, 0.849700)(0.256345, 0.836043)(0.270558, 0.822261)(0.284771, 0.808529)(0.298984, 0.794965)(0.313198, 0.781644)(0.327411, 0.768615)(0.341624, 0.755908)(0.355837, 0.743539)(0.370050, 0.731514)(0.384264, 0.719834)(0.398477, 0.708495)(0.412690, 0.697490)(0.426903, 0.686811)(0.441116, 0.676446)(0.455330, 0.666386)(0.469543, 0.656618)(0.483756, 0.647132)(0.497969, 0.637915)(0.512182, 0.628957)(0.526396, 0.620248)(0.540609, 0.611776)(0.554822, 0.603531)(0.569035, 0.595503)(0.583248, 0.587685)(0.597462, 0.580065)(0.611675, 0.572637)(0.625888, 0.565393)(0.640101, 0.558323)(0.654315, 0.551422)(0.668528, 0.544682)(0.682741, 0.538097)(0.696954, 0.531660)(0.711167, 0.525367)(0.725381, 0.519210)(0.739594, 0.513184)(0.753807, 0.507286)(0.768020, 0.501509)(0.782233, 0.495849)(0.796447, 0.490302)(0.810660, 0.484863)(0.824873, 0.479530)(0.839086, 0.474296)(0.853299, 0.469161)(0.867513, 0.464119)(0.881726, 0.459167)(0.895939, 0.454303)(0.910152, 0.449523)(0.924365, 0.444824)(0.938579, 0.440204)(0.952792, 0.435661)(0.967005, 0.431191)(0.981218, 0.426792)(0.995431, 0.422462)(1.009645, 0.418199)(1.023858, 0.414000)(1.038071, 0.409864)(1.052284, 0.405788)(1.066497, 0.401771)(1.080711, 0.397811)(1.094924, 0.393906)(1.109137, 0.390054)(1.123350, 0.386255)(1.137563, 0.382505)(1.151777, 0.378805)(1.165990, 0.375151)(1.180203, 0.371544)(1.194416, 0.367982)(1.208629, 0.364463)(1.222843, 0.360986)(1.237056, 0.357550)(1.251269, 0.354153)(1.265482, 0.350796)(1.279695, 0.347475)(1.293909, 0.344192)(1.308122, 0.340943)(1.322335, 0.337730)(1.336548, 0.334549)(1.350761, 0.331402)(1.364975, 0.328286)(1.379188, 0.325201)(1.393401, 0.322145)(1.407614, 0.319119)(1.421827, 0.316121)(1.436041, 0.313151)(1.450254, 0.310207)(1.464467, 0.307290)(1.478680, 0.304398)(1.492893, 0.301530)(1.507107, 0.298687)(1.521320, 0.295866)(1.535533, 0.293068)(1.549746, 0.290292)(1.563959, 0.287538)(1.578173, 0.284804)(1.592386, 0.282090)(1.606599, 0.279396)(1.620812, 0.276721)(1.635025, 0.274064)(1.649239, 0.271425)(1.663452, 0.268803)(1.677665, 0.266198)(1.691878, 0.263610)(1.706091, 0.261037)(1.720305, 0.258479)(1.734518, 0.255936)(1.748731, 0.253407)(1.762944, 0.250892)(1.777157, 0.248390)(1.791371, 0.245901)(1.805584, 0.243425)(1.819797, 0.240960)(1.834010, 0.238506)(1.848223, 0.236064)(1.862437, 0.233632)(1.876650, 0.231209)(1.890863, 0.228797)(1.905076, 0.226393)(1.919289, 0.223999)(1.933503, 0.221612)(1.947716, 0.219233)(1.961929, 0.216862)(1.976142, 0.214497)(1.990355, 0.212139)(2.004569, 0.209787)(2.018782, 0.207440)(2.032995, 0.205098)(2.047208, 0.202761)(2.061421, 0.200428)(2.075635, 0.198098)(2.089848, 0.195772)(2.104061, 0.193448)(2.118274, 0.191127)(2.132487, 0.188807)(2.146701, 0.186488)(2.160914, 0.184170)(2.175127, 0.181852)(2.189340, 0.179533)(2.203553, 0.177213)(2.217767, 0.174892)(2.231980, 0.172568)(2.246193, 0.170242)(2.260406, 0.167911)(2.274619, 0.165577)(2.288833, 0.163238)(2.303046, 0.160893)(2.317259, 0.158541)(2.331472, 0.156182)(2.345685, 0.153816)(2.359899, 0.151440)(2.374112, 0.149055)(2.388325, 0.146658)(2.402538, 0.144250)(2.416752, 0.141829)(2.430965, 0.139394)(2.445178, 0.136944)(2.459391, 0.134477)(2.473604, 0.131992)(2.487818, 0.129487)(2.502031, 0.126962)(2.516244, 0.124413)(2.530457, 0.121840)(2.544670, 0.119240)(2.558884, 0.116610)(2.573097, 0.113949)(2.587310, 0.111254)(2.601523, 0.108521)(2.615736, 0.105747)(2.629950, 0.102929)(2.644163, 0.100061)(2.658376, 0.097141)(2.672589, 0.094161)(2.686802, 0.091115)(2.701016, 0.087997)(2.715229, 0.084798)(2.729442, 0.081507)(2.743655, 0.078113)(2.757868, 0.074601)(2.772082, 0.070952)(2.786295, 0.067145)(2.800508, 0.063149)(2.814721, 0.058926)(2.828934, 0.054422)(2.843148, 0.049560)(2.857361, 0.044221)(2.871574, 0.038204)(2.885787, 0.031119)(2.900000, 0.021952)(2.914214, 0.000000)
};
\addlegendentry{{Marčenko–Pastur distribution}};
\addplot+[only marks,mark=x,color=red!60!white,line width=1.5pt] coordinates{(3.9722,0)};
\addlegendentry{{ Theory: $\lambda_s$ given in \eqref{eq:lambda_s} }};
\begin{scope}
    \spy[black!50!white,size=1.8cm,circle,connect spies,magnification=5] on (6.55,0.08) in node [fill=none] at (5,1.8);
\end{scope}
\end{axis}
\end{tikzpicture}
\caption{Eigenvalue distribution of $\frac1n \X\X^\T$ for $\bmu = [1.5;\mathbf{0}_{p-1}]$, $p=512$, $n=1\,024$ and $c_1 = c_2 = 1/2$.}
\label{fig:eigvenvalue-distribution-XX}
\end{center}
\vskip -0.1in
\end{figure}
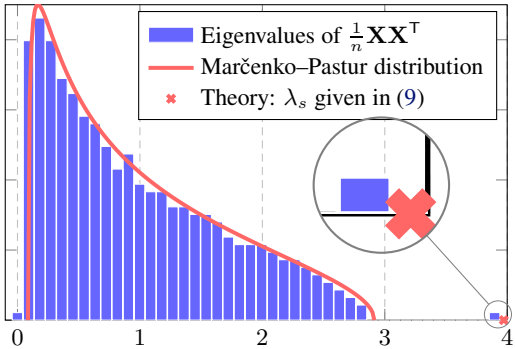

Nonetheless, the expressions in Theorem~\ref{theo:generalize-perf} and~\ref{theo:training-perf} of contour integrations are not easily analyzable nor interpretable. To gain more insight, we shall rewrite $(E, V)$ and $(E_*, V_*)$ in a more readable way. First, note from Figure~\ref{fig:eigvenvalue-distribution-XX} that the matrix $\frac1n \X\X^\T$ has (possibly) two types of eigenvalues: those inside the \emph{main bulk} (between $\lambda_- \equiv (1-\sqrt{c})^2$ and $\lambda_+ \equiv (1+\sqrt{c})^2$) of the Marčenko–Pastur distribution
\begin{equation}\label{eq:MP-distribution}
	\nu(dx) =  \frac{ \sqrt{ (x- \lambda_-)^+ (\lambda_+ - x)^+ }}{2\pi c x} dx + \left( 1- \frac1c\right)^+ \delta(x)
\end{equation}
%We defer the readers to Section~\ref{sm:intuition-eigenpair} for a more detailed discussion.
and a (possibly) isolated one\footnote{The existence (or absence) of outlying eigenvalues for the sample covariance matrix has been largely investigated in the random matrix literature and is related to the so-called ``spiked random matrix model''. We refer the reader to \cite{benaych2011eigenvalues} for an introduction. The information carried by these ``isolated'' eigenpairs also marks an important technical difference to \cite{advani2017high} in which $\X$ is only composed of noise terms.} lying away from $[\lambda_-,\lambda_+]$, that shall be treated separately. We rewrite the path $\gamma$ (that contains all eigenvalues of $\frac1n \X\X^\T$) as the sum of two paths $\gamma_b$ and $\gamma_s$, that circle around the main bulk and the isolated eigenvalue (if any), respectively. To handle the first integral of $\gamma_b$, we use the fact that for any nonzero $\lambda \in \mathbb{R}$, the limit $\lim_{z\in\mathbb{Z} \to \lambda} m(z) \equiv \check m(\lambda)$ exists \cite{silverstein1995analysis} and follow the idea in \cite{bai2008clt} by choosing the contour $\gamma_b$ to be a rectangle with sides parallel to the axes, intersecting the real axis at $0$ and $\lambda_+$ and the horizontal sides being a distance $\varepsilon \to 0$ away from the real axis, to split the contour integral into four single ones of $\check m(x)$. The second integral circling around $\gamma_s$ can be computed with the residue theorem. This together leads to the expressions of $(E, V)$ and $(E_*, V_*)$ as follows\footnote{We defer the readers to Section~\ref{sm:detailed-deduction} in Supplementary Material for a detailed exposition of Theorem~\ref{theo:generalize-perf}~and~\ref{theo:training-perf}, as well as \eqref{eq:E}-\eqref{eq:Var-star}.}
%in \eqref{eq:E}-\eqref{eq:Var-star} at the top of the next page
\begin{align}
	E &= \int \frac{ 1-f_t(x) }{x} \mu(dx) \label{eq:E}\\
	%
	%V &= \frac{\|\bmu\|^2 + c}{\|\bmu\|^2} \int \frac{ (1-f_t(x))^2 }{x^2} \mu(dx) \nonumber \\ 
	%&+ \sigma^2 \int f_t^2(x) h(dx) \label{eq:Var} \\
	V &= \frac{\|\bmu\|^2 + c}{\|\bmu\|^2} \int \frac{ (1-f_t(x))^2 \mu(dx)}{x^2}  + \sigma^2 \int f_t^2(x) \nu(dx) \label{eq:Var} \\
	E_* &= \frac{\|\bmu\|^2 + c}{\|\bmu\|^2} \int \frac{ 1-f_t(x) }{x} \mu(dx) \label{eq:E-star} \\
	%
	%V_* &= \frac{\|\bmu\|^2 + c}{\|\bmu\|^2} \int \frac{ (1-f_t(x))^2 }{x} \mu(dx) \nonumber \\ 
	%&+ \sigma^2 \int x f_t^2(x) h(dx) \label{eq:Var-star}
	V_* &= \frac{\|\bmu\|^2 + c}{\|\bmu\|^2} \int \frac{ (1-f_t(x))^2 \mu(dx)}{x} + \sigma^2 \int x f_t^2(x) \nu(dx) \label{eq:Var-star}
\end{align}
where we recall $f_t(x) = \exp(-\alpha t x)$, $\nu(x)$ given by \eqref{eq:MP-distribution} and denote the measure
\begin{equation}\label{eq:definition-measure}
	\mu(dx) \equiv \frac{\sqrt{(x-\lambda_-)^+(\lambda_+ -x)^+}}{ 2\pi(\lambda_s - x) } dx + \frac{ (\|\bmu\|^4-c)^+}{\|\bmu\|^2} \delta_{\lambda_s}(x)
\end{equation}
% attention: should be 2\pi(\lambda_s - x)^+ in the denominator
as well as
\begin{equation}\label{eq:lambda_s}
	\lambda_s = c+1 +\| \bmu\|^2 + \frac{c}{\| \bmu\|^2}\ge (\sqrt{c}+1)^2
\end{equation}
with equality if and only if $\| \bmu\|^2 = \sqrt{c}$.

A first remark on the expressions of \eqref{eq:E}-\eqref{eq:Var-star} is that $E_*$ differs from $E$ only by a factor of $\frac{\|\bmu\|^2+c}{\|\bmu\|^2}$. Also, both $V$ and $V_*$ are the sum of two parts: the first part that strongly depends on $\bmu$ and the second one that is independent of $\bmu$. One thus deduces for $\|\bmu\| \to 0$ that $E \to 0$ and
\[
	V \to \int \frac{ (1-f_t(x))^2 }{x^2} \rho(dx) + \sigma^2 \int f_t^2(x) \nu(dx) > 0
\]
% \begin{align*}
% 	V &\to \int \frac{ (1-f_t(x))^2 }{x^2} \rho(dx) + \sigma^2 \int f_t^2(x) \nu(dx) > 0\\
% 	E_* &\to \int \frac{1-f_t(x)}{x} \rho(dx) >0 \\
% 	V_* &\to \int \frac{ (1-f_t(x))^2 }{x} \rho(dx) + \sigma^2 \int xf_t^2(x) \nu(dx) > 0
% \end{align*}
with $\rho(dx) \equiv \frac{\sqrt{(x-\lambda_-)^+(\lambda_+ -x)^+}}{ 2\pi (c+1) } dx$ and therefore the generalization performance goes to $Q(0) = 0.5$. On the other hand, for $\|\bmu\| \to \infty$, one has $ \frac{E}{\sqrt{V}} \to \infty $ and hence the classifier makes perfect predictions.

In a more general context (i.e., for Gaussian mixture models with generic means and covariances as investigated in \cite{benaych2016spectral}, and obviously for practical datasets), there may be more than one eigenvalue of $\frac1n \X\X^\T$ lying outside the main bulk, which may not be limited to the interval $[\lambda_-,\lambda_+]$. In this case, the expression of $m(z)$, instead of being explicitly given by \eqref{eq:MP-equation}, may be determined through more elaborate (often implicit) formulations. While handling more generic models is technically reachable within the present analysis scheme, the results are much less intuitive. Similar objectives cannot be achieved within the framework presented in \cite{advani2017high}; this conveys more practical interest to our results and the proposed analysis framework.
%Our analysis scheme can be easily extended to handle these more intricate cases.

%%% eigenvalue distribution and MP law

\section{Discussions}
\label{sec:discuss}

In this section, with a careful inspection of \eqref{eq:E} and~\eqref{eq:Var}, discussions will be made from several different aspects. First of all, recall that the generalization performance is simply given by $ Q\left( \frac{\bmu^\T \w(t)}{ \| \w(t) \| } \right)$, with the term $\frac{\bmu^\T \w(t)}{ \| \w(t) \| }$ describing the alignment between $\w(t)$ and $\bmu$, therefore the best possible generalization performance is simply $Q(\|\bmu\|)$. Nonetheless, this ``best'' performance can never be achieved as long as $p/n \to c >0$, as described in the following remark.

\begin{Remark}[Optimal Generalization Performance]
Note that, with Cauchy–Schwarz inequality and the fact that $\int \mu(dx) = \| \bmu \|^2$ from \eqref{eq:definition-measure}, one has
%1_{x\in[\lambda_-,\lambda_+]}
\[
	E^2 \le \int \frac{(1-f_t(x))^2}{x^2} d\mu(x) \cdot \int d\mu(x) \le \frac{\|\bmu\|^4}{\|\bmu\|^2+c} V
\]
with equality in the right-most inequality if and only if the variance $\sigma^2 = 0$. One thus concludes that $E/\sqrt{V} \le \|\bmu\|^2/\sqrt{\|\bmu\|^2 + c}$ and the best generalization performance (lowest misclassification rate) is $Q (\|\bmu\|^2/\sqrt{\|\bmu\|^2 + c})$ and can be attained only when $\sigma^2 = 0$.
\label{rem:optimal-generalization-perf}
\end{Remark}

The above remark is of particular interest because, for a given task (thus $p, \bmu$ fixed) it allows one to compute the \emph{minimum} training data number $n$ to fulfill a certain request of classification accuracy.

As a side remark, note that in the expression of $E/\sqrt{V}$ the initialization variance $\sigma^2$ only appears in $V$, meaning that random initializations impair the generalization performance of the network. As such, one should initialize with $\sigma^2$ very close, but not equal, to zero, to obtain symmetry breaking between hidden units \cite{goodfellow2016deeplearning} as well as to mitigate the drop of performance due to large $\sigma^2$.

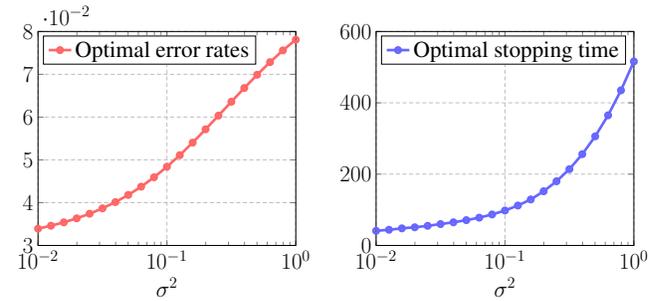
\begin{figure}[tbh]
\vskip 0.1in
\begin{center}
\begin{minipage}[b]{0.48\columnwidth}%
%\begin{tikzpicture}[font=\Large,scale=0.55]
\begin{tikzpicture}[font=\LARGE,scale=0.5]
\renewcommand{\axisdefaulttryminticks}{4} 
\pgfplotsset{every major grid/.style={densely dashed}}       
\tikzstyle{every axis y label}+=[yshift=-10pt] 
\tikzstyle{every axis x label}+=[yshift=5pt]
\pgfplotsset{every axis legend/.append style={cells={anchor=west},fill=white, at={(0.02,0.98)}, anchor=north west, font=\LARGE }}
\begin{axis}[
xmode=log,
xmin=0.01,
ymin=0.03,
xmax=1,
ymax=0.08,
xlabel={$\sigma^2$},
grid=major,
scaled ticks=true,
]
\addplot[mark=o,color=red!60!white,line width=2pt] coordinates{
(0.010000,0.033928)(0.012589,0.034610)(0.015849,0.035407)(0.019953,0.036334)(0.025119,0.037416)(0.031623,0.038674)(0.039811,0.040131)(0.050119,0.041812)(0.063096,0.043740)(0.079433,0.045930)(0.100000,0.048388)(0.125893,0.051103)(0.158489,0.054040)(0.199526,0.057145)(0.251189,0.060351)(0.316228,0.063587)(0.398107,0.066787)(0.501187,0.069890)(0.630957,0.072840)(0.794328,0.075589)(1.000000,0.078101)
};
\addlegendentry{{Optimal error rates}};
\end{axis}
\end{tikzpicture}
\end{minipage}%
\hfill{}
\begin{minipage}[b]{0.48\columnwidth}%
\begin{tikzpicture}[font=\LARGE,scale=0.5]
\renewcommand{\axisdefaulttryminticks}{4} 
\pgfplotsset{every major grid/.style={densely dashed}}       
\tikzstyle{every axis y label}+=[yshift=-10pt] 
\tikzstyle{every axis x label}+=[yshift=5pt]
\pgfplotsset{every axis legend/.append style={cells={anchor=west},fill=white, at={(0.02,0.98)}, anchor=north west, font=\LARGE }}
\begin{axis}[
xmode=log,
xmin=0.01,
ymin=0,
xmax=1,
ymax=600,
xlabel={$\sigma^2$},
%xtick scale label={100},
%ylabel={Misclassification rate},
%ylabel={Integral value},
% ytick={0,0.05,0.1,0.15,0.2},
% yticklabels={$0$,$0.05$,$0.1$,$0.15$,$0.2$},
grid=major,
scaled ticks=true,
]
\addplot[mark=o,color=blue!60!white,line width=2pt] coordinates{
(0.010000,41.000000)(0.012589,44.000000)(0.015849,48.000000)(0.019953,51.000000)(0.025119,55.000000)(0.031623,60.000000)(0.039811,65.000000)(0.050119,71.000000)(0.063096,78.000000)(0.079433,87.000000)(0.100000,98.000000)(0.125893,112.000000)(0.158489,129.000000)(0.199526,152.000000)(0.251189,180.000000)(0.316228,214.000000)(0.398107,256.000000)(0.501187,306.000000)(0.630957,365.000000)(0.794328,435.000000)(1.000000,516.000000)
};
\addlegendentry{{Optimal stopping time}};
\end{axis}
\end{tikzpicture}
\end{minipage}
\caption{ Optimal performance and corresponding stopping time as functions of $\sigma^2$, with $c = 1/2$, $\|\bmu\|^2=4$ and $\alpha=0.01$.}
\label{fig:optimal-perf-and-time-vs-sigma2}
\end{center}
\vskip -0.1in
\end{figure}

In Figure~\ref{fig:optimal-perf-and-time-vs-sigma2} we plot the optimal generalization performance with the corresponding optimal stopping time as functions of $\sigma^2$, showing that small initialization helps training in terms of both accuracy and efficiency.

%%% Optimal perf and stopping time versus sigma2

Although the integrals in \eqref{eq:E} and \eqref{eq:Var} do not have nice closed forms, note that, for $t$ close to $0$, with a Taylor expansion of $f_t(x) \equiv \exp(-\alpha tx)$ around $\alpha t x = 0$, one gets more interpretable forms of $E$ and $V$ without integrals, as presented in the following subsection.

\subsection{Approximation for $t$ close to $0$}
\label{subsec:t-close-to-0}

Taking $t = 0$, one has $f_t(x) = 1$ and therefore $E = 0$, $V= \sigma^2 \int \nu(dx)= \sigma^2$, with $\nu(dx)$ the Marčenko–Pastur distribution given in \eqref{eq:MP-distribution}. As a consequence, at the beginning stage of training, the generalization performance is $Q(0) = 0.5$ for $\sigma^2 \neq 0$ and the classifier makes random guesses.

For $t$ not equal but close to $0$, the Taylor expansion of $f_t(x) \equiv \exp(-\alpha tx)$ around $\alpha t x=0$ gives 
\[
 	f_t(x) \equiv \exp(-\alpha t x) \approx 1 -\alpha t x + O(\alpha^2 t^2 x^2).
\]

Making the substitution $x = 1+c-2\sqrt{c} \cos\theta$ and with the fact that $\int_0^{\pi} \frac{ \sin^2\theta }{ p + q \cos\theta } d\theta = \frac{p \pi}{q^2} \left( 1 - \sqrt{1-q^2/p^2 } \right)$ (see for example 3.644-5 in \cite{gradshteyn2014table}), one gets $E = \tilde{E} + O(\alpha^2 t^2)$ and $V = \tilde{V}+ O(\alpha^2 t^2)$, where
\[
	\tilde{E} \equiv \frac{\alpha t}{2} g(\bmu,c) + \frac{ (\|\bmu \|^4 -c)^+ }{\| \bmu \|^2} \alpha t = \| \bmu \|^2 \alpha t
\]
\begin{align*}
	\tilde{V} &\equiv \frac{\|\bmu\|^2 + c}{\|\bmu\|^2} \frac{ (\|\bmu \|^4 - c)^+}{\| \bmu \|^2} \alpha^2 t^2 + \frac{\|\bmu\|^2 + c}{\|\bmu\|^2} \frac{\alpha^2 t^2}2  g(\bmu,c) \\
	%%%
	& + \sigma^2 (1+c) \alpha^2 t^2 - 2\sigma^2 \alpha t + \left(1-\frac1c\right)^+ \sigma^2 \\
	%%%
	&+ \frac{\sigma^2}{2c} \left( 1+c - (1+\sqrt{c}) |1-\sqrt{c}| \right)  \\
	%%% 
	&= (\| \bmu\|^2 + c + c\sigma^2) \alpha^2 t^2 + \sigma^2 ( \alpha t - 1 )^2 
\end{align*}
with $g(\bmu,c) \equiv \| \bmu\|^2 + \frac{c}{\| \bmu\|^2} - \left( \| \bmu\| + \frac{\sqrt{c}}{\| \bmu\|} \right) \left| \| \bmu\| - \frac{ \sqrt{c} }{\| \bmu\|} \right| $ and consequently $\frac12 g(\bmu,c) + \frac{(\|\bmu \|^4 -c)^+}{\| \bmu \|^2} = \| \bmu\|^2$. It is interesting to note from the above calculation that, although $E$ and $V$ seem to have different behaviors\footnote{This phenomenon has been largely observed in random matrix theory and is referred to as ``phase transition''\cite{baik2005phase}.} for $\| \bmu\|^2 > \sqrt{c}$ or $c>1$, it is in fact not the case and the extra part of $\| \bmu\|^2 > \sqrt{c}$ (or $c>1$) compensates for the singularity of the integral, so that the generalization performance of the classifier is a smooth function of both $\| \bmu\|^2 $ and $c$.

Taking the derivative of $\frac{ \tilde{E}}{ \sqrt{ \tilde{V}} } $ with respect to $t$, one has
\[
	\frac{\partial }{\partial t} \frac{ \tilde{E}}{ \sqrt{ \tilde{V}} } = \frac{ \alpha (1-\alpha t) \sigma^2 }{ {\tilde V}^{3/2} }
\]
which implies that the maximum of $\frac{ \tilde{E}} {\sqrt{ \tilde{V}}}$ is  $ \frac{ \| \bmu\|^2 }{ \sqrt{ \| \bmu\|^2 +c+c\sigma^2} }$ and can be attained with $t = 1/\alpha$. Moreover, taking $t=0$ in the above equation one gets $\frac{\partial }{\partial t} \frac{ \tilde{E}}{ \sqrt{ \tilde{V}} } \big|_{t=0} = \frac{\alpha}{\sigma}$. Therefore, large $\sigma$ is harmful to the training efficiency, which coincides with the conclusion from Remark~\ref{rem:optimal-generalization-perf}.

%\textbf{can we bound the true error by this approximation? Maybe YES! But the optimal stopping time make no sense in this case...}

The approximation error arising from Taylor expansion can be large for $t$ away from $0$, e.g., at $t = 1/\alpha$ the difference $E - \tilde{E}$ is of order $O(1)$ and thus cannot be neglected. 

%Nonetheless, some insights from the above calculations contribute to the understanding of the object of interest $\frac{E}{ \sqrt{V} }$, as presented in the following remark.

\subsection{As $t \to \infty$: least-squares solution}
As $t \to \infty$, one has $f_t(x) \to 0$ which results in the least-square solution $\w_{LS} = (\X\X^\T)^{-1} \X \y $ or $\w_{LS} = \X (\X^\T\X)^{-1} \y $ and consequently
\begin{equation}
	\frac{ \bmu^\T \w_{LS} }{ \| \w_{LS}\| } = \frac{ \| \bmu \|^2 }{\sqrt{\| \bmu \|^2 + c}}  \sqrt{ 
	1-\min\left(c,\frac1c\right) }.
	\label{eq:perf-w-LS}
\end{equation}

Comparing \eqref{eq:perf-w-LS} with the expression in Remark~\ref{rem:optimal-generalization-perf}, one observes that when $t \to \infty$ the network becomes ``over-trained'' and the performance drops by a factor of $\sqrt{1- \min (c, c^{-1} ) }$. This becomes even worse when $c$ gets close to $1$, as is consistent with the empirical findings in \cite{advani2017high}. However, the point $c=1$ is a singularity for \eqref{eq:perf-w-LS}, but not for $\frac{E}{ \sqrt{V} }$ as in \eqref{eq:E} and~\eqref{eq:Var}. One may thus expect to have a smooth and reliable behavior of the well-trained network for $c$ close to $1$, which is a noticeable advantage of gradient-based training compared to simple least-square method. This coincides with the conclusion of \cite{yao2007early} in which the asymptotic behavior of solely $n \to \infty$ is considered.

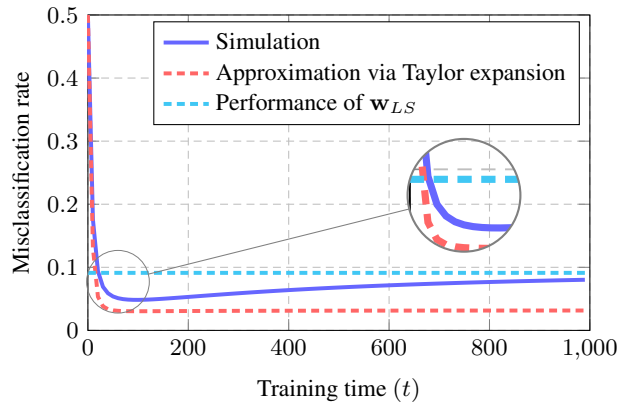
\begin{figure}[htb]
\vskip 0.1in
\begin{center}
\begin{tikzpicture}[font=\footnotesize,spy using outlines]
\renewcommand{\axisdefaulttryminticks}{4} 
\pgfplotsset{every major grid/.style={densely dashed}}       
\tikzstyle{every axis y label}+=[yshift=-10pt] 
\tikzstyle{every axis x label}+=[yshift=5pt]
\pgfplotsset{every axis legend/.append style={cells={anchor=west},fill=white, at={(0.98,0.98)}, anchor=north east, font=\footnotesize }}
\begin{axis}[
width=\columnwidth,
height=0.7\columnwidth,
xmin=0,
xmax=1000,
ymin=0,
ymax=.5,
grid=major,
xlabel={Training time $(t)$},
ylabel={Misclassification rate},
ytick={0,0.1,0.2,0.3,0.4,0.5},
scaled ticks=true,
]
\addplot[color=blue!60!white,line width=1.5pt] coordinates{
(0.000000,0.500000)(10.000000,0.175906)(20.000000,0.093935)(30.000000,0.069117)(40.000000,0.058867)(50.000000,0.053804)(60.000000,0.051075)(70.000000,0.049573)(80.000000,0.048786)(90.000000,0.048446)(100.000000,0.048399)(110.000000,0.048550)(120.000000,0.048839)(130.000000,0.049225)(140.000000,0.049681)(150.000000,0.050186)(160.000000,0.050726)(170.000000,0.051290)(180.000000,0.051871)(190.000000,0.052463)(200.000000,0.053060)(210.000000,0.053660)(220.000000,0.054260)(230.000000,0.054857)(240.000000,0.055450)(250.000000,0.056038)(260.000000,0.056620)(270.000000,0.057195)(280.000000,0.057762)(290.000000,0.058320)(300.000000,0.058870)(310.000000,0.059412)(320.000000,0.059944)(330.000000,0.060468)(340.000000,0.060982)(350.000000,0.061487)(360.000000,0.061983)(370.000000,0.062470)(380.000000,0.062948)(390.000000,0.063418)(400.000000,0.063878)(410.000000,0.064330)(420.000000,0.064774)(430.000000,0.065209)(440.000000,0.065636)(450.000000,0.066055)(460.000000,0.066466)(470.000000,0.066869)(480.000000,0.067264)(490.000000,0.067652)(500.000000,0.068033)(510.000000,0.068406)(520.000000,0.068773)(530.000000,0.069133)(540.000000,0.069485)(550.000000,0.069832)(560.000000,0.070171)(570.000000,0.070505)(580.000000,0.070832)(590.000000,0.071153)(600.000000,0.071469)(610.000000,0.071778)(620.000000,0.072082)(630.000000,0.072380)(640.000000,0.072673)(650.000000,0.072961)(660.000000,0.073243)(670.000000,0.073520)(680.000000,0.073793)(690.000000,0.074060)(700.000000,0.074323)(710.000000,0.074580)(720.000000,0.074834)(730.000000,0.075083)(740.000000,0.075327)(750.000000,0.075567)(760.000000,0.075803)(770.000000,0.076035)(780.000000,0.076263)(790.000000,0.076487)(800.000000,0.076707)(810.000000,0.076923)(820.000000,0.077136)(830.000000,0.077344)(840.000000,0.077550)(850.000000,0.077752)(860.000000,0.077950)(870.000000,0.078145)(880.000000,0.078337)(890.000000,0.078525)(900.000000,0.078711)(910.000000,0.078893)(920.000000,0.079072)(930.000000,0.079249)(940.000000,0.079422)(950.000000,0.079592)(960.000000,0.079760)(970.000000,0.079925)(980.000000,0.080087)(990.000000,0.080247)
};
\addlegendentry{{ Simulation }};
\addplot[densely dashed,color=red!60!white, line width=1.5pt] coordinates{
(0.000000,0.500000)(10.000000,0.130370)(20.000000,0.053377)(30.000000,0.038181)(40.000000,0.033586)(50.000000,0.031801)(60.000000,0.031011)(70.000000,0.030641)(80.000000,0.030469)(90.000000,0.030398)(100.000000,0.030381)(110.000000,0.030392)(120.000000,0.030420)(130.000000,0.030456)(140.000000,0.030496)(150.000000,0.030538)(160.000000,0.030580)(170.000000,0.030621)(180.000000,0.030661)(190.000000,0.030699)(200.000000,0.030735)(210.000000,0.030770)(220.000000,0.030803)(230.000000,0.030834)(240.000000,0.030863)(250.000000,0.030891)(260.000000,0.030918)(270.000000,0.030943)(280.000000,0.030967)(290.000000,0.030990)(300.000000,0.031011)(310.000000,0.031032)(320.000000,0.031051)(330.000000,0.031070)(340.000000,0.031088)(350.000000,0.031105)(360.000000,0.031121)(370.000000,0.031136)(380.000000,0.031151)(390.000000,0.031165)(400.000000,0.031179)(410.000000,0.031192)(420.000000,0.031204)(430.000000,0.031216)(440.000000,0.031228)(450.000000,0.031239)(460.000000,0.031250)(470.000000,0.031260)(480.000000,0.031270)(490.000000,0.031280)(500.000000,0.031289)(510.000000,0.031298)(520.000000,0.031307)(530.000000,0.031315)(540.000000,0.031323)(550.000000,0.031331)(560.000000,0.031338)(570.000000,0.031346)(580.000000,0.031353)(590.000000,0.031360)(600.000000,0.031366)(610.000000,0.031373)(620.000000,0.031379)(630.000000,0.031385)(640.000000,0.031391)(650.000000,0.031397)(660.000000,0.031403)(670.000000,0.031408)(680.000000,0.031413)(690.000000,0.031419)(700.000000,0.031424)(710.000000,0.031429)(720.000000,0.031433)(730.000000,0.031438)(740.000000,0.031443)(750.000000,0.031447)(760.000000,0.031451)(770.000000,0.031456)(780.000000,0.031460)(790.000000,0.031464)(800.000000,0.031468)(810.000000,0.031472)(820.000000,0.031475)(830.000000,0.031479)(840.000000,0.031483)(850.000000,0.031486)(860.000000,0.031489)(870.000000,0.031493)(880.000000,0.031496)(890.000000,0.031499)(900.000000,0.031503)(910.000000,0.031506)(920.000000,0.031509)(930.000000,0.031512)(940.000000,0.031515)(950.000000,0.031517)(960.000000,0.031520)(970.000000,0.031523)(980.000000,0.031526)(990.000000,0.031528)
};
\addlegendentry{{ Approximation via Taylor expansion }}; 
\addplot[densely dashed,color=cyan!60!white, line width=1.5pt] coordinates{
(0.000000,0.091211)(10.000000,0.091211)(20.000000,0.091211)(30.000000,0.091211)(40.000000,0.091211)(50.000000,0.091211)(60.000000,0.091211)(70.000000,0.091211)(80.000000,0.091211)(90.000000,0.091211)(100.000000,0.091211)(110.000000,0.091211)(120.000000,0.091211)(130.000000,0.091211)(140.000000,0.091211)(150.000000,0.091211)(160.000000,0.091211)(170.000000,0.091211)(180.000000,0.091211)(190.000000,0.091211)(200.000000,0.091211)(210.000000,0.091211)(220.000000,0.091211)(230.000000,0.091211)(240.000000,0.091211)(250.000000,0.091211)(260.000000,0.091211)(270.000000,0.091211)(280.000000,0.091211)(290.000000,0.091211)(300.000000,0.091211)(310.000000,0.091211)(320.000000,0.091211)(330.000000,0.091211)(340.000000,0.091211)(350.000000,0.091211)(360.000000,0.091211)(370.000000,0.091211)(380.000000,0.091211)(390.000000,0.091211)(400.000000,0.091211)(410.000000,0.091211)(420.000000,0.091211)(430.000000,0.091211)(440.000000,0.091211)(450.000000,0.091211)(460.000000,0.091211)(470.000000,0.091211)(480.000000,0.091211)(490.000000,0.091211)(500.000000,0.091211)(510.000000,0.091211)(520.000000,0.091211)(530.000000,0.091211)(540.000000,0.091211)(550.000000,0.091211)(560.000000,0.091211)(570.000000,0.091211)(580.000000,0.091211)(590.000000,0.091211)(600.000000,0.091211)(610.000000,0.091211)(620.000000,0.091211)(630.000000,0.091211)(640.000000,0.091211)(650.000000,0.091211)(660.000000,0.091211)(670.000000,0.091211)(680.000000,0.091211)(690.000000,0.091211)(700.000000,0.091211)(710.000000,0.091211)(720.000000,0.091211)(730.000000,0.091211)(740.000000,0.091211)(750.000000,0.091211)(760.000000,0.091211)(770.000000,0.091211)(780.000000,0.091211)(790.000000,0.091211)(800.000000,0.091211)(810.000000,0.091211)(820.000000,0.091211)(830.000000,0.091211)(840.000000,0.091211)(850.000000,0.091211)(860.000000,0.091211)(870.000000,0.091211)(880.000000,0.091211)(890.000000,0.091211)(900.000000,0.091211)(910.000000,0.091211)(920.000000,0.091211)(930.000000,0.091211)(940.000000,0.091211)(950.000000,0.091211)(960.000000,0.091211)(970.000000,0.091211)(980.000000,0.091211)(990.000000,0.091211)
};
\addlegendentry{{ Performance of $\w_{LS}$ }}; 
%
% \addplot+[only marks,mark=x,color=cyan!60!white,line width=1pt] coordinates{ (100.000000,0.048399)(100.000000,0.030381)  };
% \addlegendentry{{ Performance at $t = 1/\alpha$ }}; 
\begin{scope}
    \spy[black!50!white,size=1.5cm,circle,connect spies,magnification=1.8] on (0.4,0.65) in node [fill=none] at (5,1.8);
\end{scope}
\end{axis}
\end{tikzpicture}
\caption{ Generalization performance for $\bmu = \begin{bmatrix}2;\mathbf{0}_{p-1}\end{bmatrix}$, $p=256$, $n=512$, $c_1 = c_2 = 1/2$, $\sigma^2 = 0.1$ and $\alpha = 0.01$. Simulation results obtained by averaging over $50$ runs.}
\label{fig:approximation-t-small}
\end{center}
\vskip -0.1in
\end{figure}

In Figure~\ref{fig:approximation-t-small} we plot the generalization performance from simulation (blue line),  the approximation from Taylor expansion of $f_t(x)$ as described in Section~\ref{subsec:t-close-to-0} (red dashed line), together with the performance of $\w_{LS}$ (cyan dashed line). One observes a close match between the result from Taylor expansion and the true performance for $t$ small, with the former being optimal at $t=100$ and the latter slowly approaching the performance of $\w_{LS}$ as $t$ goes to infinity.
%the theoretical optimal performance in Remark~\ref{rem:optimal-generalization-perf} (cyan line),

\begin{figure}[htb]
\vskip 0.1in
\begin{center}
\begin{tikzpicture}[font=\footnotesize,spy using outlines]
\renewcommand{\axisdefaulttryminticks}{4} 
\pgfplotsset{every major grid/.style={densely dashed}}       
\tikzstyle{every axis y label}+=[yshift=-10pt] 
\tikzstyle{every axis x label}+=[yshift=5pt]
\pgfplotsset{every axis legend/.append style={cells={anchor=west},fill=white, at={(0.98,0.95)}, anchor=north east, font=\footnotesize }}
\begin{axis}[
width=\columnwidth,
height=0.7\columnwidth,
xmin=0,
xmax=1000,
ymin=0.01,
ymax=0.51,
grid=major,
xlabel={Training time $(t)$},
ylabel={Misclassification rate},
ytick={0,0.1,0.2,0.3,0.4,0.5},
scaled ticks=true,
]
\addplot[color=blue!60!white,line width=1.5pt] coordinates{
(0.000000,0.488750)(10.000000,0.177031)(20.000000,0.103750)(30.000000,0.079922)(40.000000,0.071797)(50.000000,0.067656)(60.000000,0.066562)(70.000000,0.065625)(80.000000,0.066719)(90.000000,0.066875)(100.000000,0.067969)(110.000000,0.069375)(120.000000,0.070391)(130.000000,0.071719)(140.000000,0.072500)(150.000000,0.074609)(160.000000,0.075078)(170.000000,0.077109)(180.000000,0.077969)(190.000000,0.078516)(200.000000,0.079766)(210.000000,0.081016)(220.000000,0.082578)(230.000000,0.083594)(240.000000,0.084375)(250.000000,0.085234)(260.000000,0.086250)(270.000000,0.087891)(280.000000,0.089453)(290.000000,0.090391)(300.000000,0.091016)(310.000000,0.092109)(320.000000,0.093203)(330.000000,0.093906)(340.000000,0.094766)(350.000000,0.095469)(360.000000,0.096250)(370.000000,0.097500)(380.000000,0.098047)(390.000000,0.098672)(400.000000,0.099844)(410.000000,0.100781)(420.000000,0.101562)(430.000000,0.102344)(440.000000,0.103438)(450.000000,0.104297)(460.000000,0.104609)(470.000000,0.105391)(480.000000,0.105547)(490.000000,0.106016)(500.000000,0.106875)(510.000000,0.107578)(520.000000,0.108281)(530.000000,0.108828)(540.000000,0.109609)(550.000000,0.109922)(560.000000,0.110234)(570.000000,0.111172)(580.000000,0.111797)(590.000000,0.112422)(600.000000,0.112500)(610.000000,0.112812)(620.000000,0.113672)(630.000000,0.114062)(640.000000,0.114375)(650.000000,0.114922)(660.000000,0.115547)(670.000000,0.115937)(680.000000,0.116562)(690.000000,0.117031)(700.000000,0.117656)(710.000000,0.117813)(720.000000,0.118594)(730.000000,0.119766)(740.000000,0.120469)(750.000000,0.120703)(760.000000,0.121016)(770.000000,0.121406)(780.000000,0.122031)(790.000000,0.122656)(800.000000,0.122969)(810.000000,0.123281)(820.000000,0.123906)(830.000000,0.124297)(840.000000,0.124688)(850.000000,0.125469)(860.000000,0.125938)(870.000000,0.126328)(880.000000,0.126953)(890.000000,0.127812)(900.000000,0.128125)(910.000000,0.128594)(920.000000,0.129297)(930.000000,0.130078)(940.000000,0.130547)(950.000000,0.131094)(960.000000,0.131641)(970.000000,0.132109)(980.000000,0.132812)(990.000000,0.133125)
};
\addlegendentry{{ Simulation }};
\addplot[densely dashed,color=red!60!white, line width=1.5pt] coordinates{
(0.000000,0.500000)(10.000000,0.135456)(20.000000,0.061133)(30.000000,0.046126)(40.000000,0.041512)(50.000000,0.039705)(60.000000,0.038903)(70.000000,0.038526)(80.000000,0.038351)(90.000000,0.038279)(100.000000,0.038261)(110.000000,0.038273)(120.000000,0.038301)(130.000000,0.038338)(140.000000,0.038379)(150.000000,0.038422)(160.000000,0.038464)(170.000000,0.038506)(180.000000,0.038546)(190.000000,0.038585)(200.000000,0.038622)(210.000000,0.038657)(220.000000,0.038691)(230.000000,0.038722)(240.000000,0.038752)(250.000000,0.038781)(260.000000,0.038808)(270.000000,0.038834)(280.000000,0.038858)(290.000000,0.038881)(300.000000,0.038903)(310.000000,0.038924)(320.000000,0.038944)(330.000000,0.038963)(340.000000,0.038981)(350.000000,0.038998)(360.000000,0.039014)(370.000000,0.039030)(380.000000,0.039045)(390.000000,0.039060)(400.000000,0.039073)(410.000000,0.039087)(420.000000,0.039099)(430.000000,0.039112)(440.000000,0.039123)(450.000000,0.039135)(460.000000,0.039146)(470.000000,0.039156)(480.000000,0.039166)(490.000000,0.039176)(500.000000,0.039185)(510.000000,0.039194)(520.000000,0.039203)(530.000000,0.039212)(540.000000,0.039220)(550.000000,0.039228)(560.000000,0.039235)(570.000000,0.039243)(580.000000,0.039250)(590.000000,0.039257)(600.000000,0.039264)(610.000000,0.039271)(620.000000,0.039277)(630.000000,0.039283)(640.000000,0.039289)(650.000000,0.039295)(660.000000,0.039301)(670.000000,0.039306)(680.000000,0.039312)(690.000000,0.039317)(700.000000,0.039322)(710.000000,0.039327)(720.000000,0.039332)(730.000000,0.039337)(740.000000,0.039341)(750.000000,0.039346)(760.000000,0.039350)(770.000000,0.039354)(780.000000,0.039359)(790.000000,0.039363)(800.000000,0.039367)(810.000000,0.039371)(820.000000,0.039374)(830.000000,0.039378)(840.000000,0.039382)(850.000000,0.039385)(860.000000,0.039389)(870.000000,0.039392)(880.000000,0.039396)(890.000000,0.039399)(900.000000,0.039402)(910.000000,0.039405)(920.000000,0.039408)(930.000000,0.039411)(940.000000,0.039414)(950.000000,0.039417)(960.000000,0.039420)(970.000000,0.039423)(980.000000,0.039426)(990.000000,0.039428)
};
\addlegendentry{{ Approximation via Taylor expansion }}; 
\addplot[densely dashed,color=cyan!60!white, line width=1.5pt] coordinates{
(0.000000,0.500000)(10.000000,0.500000)(20.000000,0.500000)(30.000000,0.500000)(40.000000,0.500000)(50.000000,0.500000)(60.000000,0.500000)(70.000000,0.500000)(80.000000,0.500000)(90.000000,0.500000)(100.000000,0.500000)(110.000000,0.500000)(120.000000,0.500000)(130.000000,0.500000)(140.000000,0.500000)(150.000000,0.500000)(160.000000,0.500000)(170.000000,0.500000)(180.000000,0.500000)(190.000000,0.500000)(200.000000,0.500000)(210.000000,0.500000)(220.000000,0.500000)(230.000000,0.500000)(240.000000,0.500000)(250.000000,0.500000)(260.000000,0.500000)(270.000000,0.500000)(280.000000,0.500000)(290.000000,0.500000)(300.000000,0.500000)(310.000000,0.500000)(320.000000,0.500000)(330.000000,0.500000)(340.000000,0.500000)(350.000000,0.500000)(360.000000,0.500000)(370.000000,0.500000)(380.000000,0.500000)(390.000000,0.500000)(400.000000,0.500000)(410.000000,0.500000)(420.000000,0.500000)(430.000000,0.500000)(440.000000,0.500000)(450.000000,0.500000)(460.000000,0.500000)(470.000000,0.500000)(480.000000,0.500000)(490.000000,0.500000)(500.000000,0.500000)(510.000000,0.500000)(520.000000,0.500000)(530.000000,0.500000)(540.000000,0.500000)(550.000000,0.500000)(560.000000,0.500000)(570.000000,0.500000)(580.000000,0.500000)(590.000000,0.500000)(600.000000,0.500000)(610.000000,0.500000)(620.000000,0.500000)(630.000000,0.500000)(640.000000,0.500000)(650.000000,0.500000)(660.000000,0.500000)(670.000000,0.500000)(680.000000,0.500000)(690.000000,0.500000)(700.000000,0.500000)(710.000000,0.500000)(720.000000,0.500000)(730.000000,0.500000)(740.000000,0.500000)(750.000000,0.500000)(760.000000,0.500000)(770.000000,0.500000)(780.000000,0.500000)(790.000000,0.500000)(800.000000,0.500000)(810.000000,0.500000)(820.000000,0.500000)(830.000000,0.500000)(840.000000,0.500000)(850.000000,0.500000)(860.000000,0.500000)(870.000000,0.500000)(880.000000,0.500000)(890.000000,0.500000)(900.000000,0.500000)(910.000000,0.500000)(920.000000,0.500000)(930.000000,0.500000)(940.000000,0.500000)(950.000000,0.500000)(960.000000,0.500000)(970.000000,0.500000)(980.000000,0.500000)(990.000000,0.500000)
};
\addlegendentry{{ Performance of $\w_{LS}$ }}; 
% \begin{scope}
%     \spy[black!50!white,size=1.5cm,circle,connect spies,magnification=1.8] on (0.5,0.5) in node [fill=none] at (5,1.25);
% \end{scope}
\end{axis}
\end{tikzpicture}
\caption{ Generalization performance for $\bmu = \begin{bmatrix}2;\mathbf{0}_{p-1}\end{bmatrix}$, $p=512$, $n=512$, $c_1 = c_2 = 1/2$, $\sigma^2 = 0.1$ and $\alpha = 0.01$. Simulation results obtained by averaging over $50$ runs.}
\label{fig:approximation-c=1}
\end{center}
\vskip -0.1in
\end{figure}
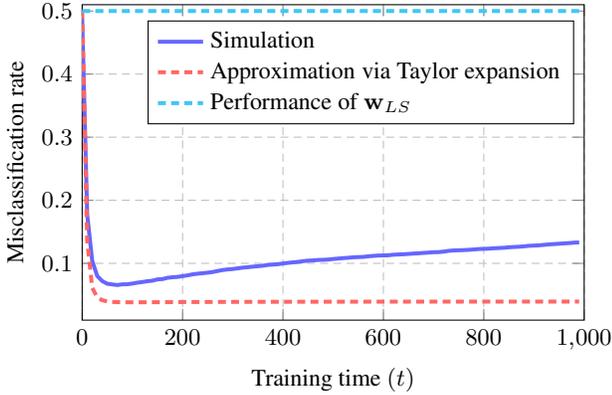

In Figure~\ref{fig:approximation-c=1} we underline the case $c=1$ by taking $p=n=512$ with all other parameters unchanged from Figure~\ref{fig:approximation-t-small}. One observes that the simulation curve (blue line) increases much faster compared to Figure~\ref{fig:approximation-t-small} and is supposed to end up at $0.5$, which is the performance of $\w_{LS}$ (cyan dashed line). This confirms a serious degradation of performance for $c$ close to $1$ of the classical least-squares solution. 

\subsection{Special case for $c = 0$}
\label{subsec:c}

One major interest of random matrix analysis is that the ratio $c$ appears constantly in the analysis. Taking $c = 0$ signifies that we have far more training data than their dimension. This results in both $\lambda_-$, $\lambda_+ \to 1$, $\lambda_s \to 1 + \| \bmu \|^2$ and
\begin{align*}
	E &\to \| \bmu \|^2 \frac{ 1-f_t(1 + \| \bmu \|^2 ) }{ 1+ \| \bmu \|^2 } \\
	V &\to \| \bmu \|^2  \left( \frac{ 1-f_t(1 + \| \bmu \|^2 ) }{ 1+ \| \bmu \|^2 } \right)^2 + \sigma^2 f_t^2(1).
\end{align*}

As a consequence, $\frac{ E }{ \sqrt{ V} } \to \| \bmu\|$ if $\sigma^2 = 0$. This can be explained by the fact that with sufficient training data the classifier learns to align perfectly to $\bmu$ so that $\frac{ \bmu^\T \w(t) }{ \|\w(t) \| } = \| \bmu \|$. On the other hand, with initialization $\sigma^2 \neq 0$, one always has $\frac{ E }{ \sqrt{ V} } < \| \bmu\|$. But still, as $t$ goes large, the network forgets the initialization exponentially fast and converges to the optimal $\w(t)$ that aligns to $\bmu$. 

In particular, for $\sigma^2 \neq 0$, we are interested in the optimal stopping time by taking the derivative with respect to $t$,
\[
	\frac{\partial }{\partial t} \frac{ E }{ \sqrt{ V } } = \frac{\alpha \sigma^2 \|\bmu\|^2}{ V^{3/2} } \frac{ \|\bmu\|^2 f_t(1+\|\bmu\|^2) + 1 }{1+\|\bmu\|^2} f_t^2(1) > 0
\]
showing that when $c= 0$, the generalization performance continues to increase as $t$ grows and there is in fact no ``over-training'' in this case.
\begin{figure}[htb]
\vskip 0.1in
\begin{center}
\begin{tikzpicture}[font=\footnotesize,spy using outlines]
\renewcommand{\axisdefaulttryminticks}{4} 
\pgfplotsset{every major grid/.style={densely dashed}}       
\tikzstyle{every axis y label}+=[yshift=-10pt] 
\tikzstyle{every axis x label}+=[yshift=5pt]
\pgfplotsset{every axis legend/.append style={cells={anchor=west},fill=white, at={(0.98,0.98)}, anchor=north east, font=\footnotesize }}
\begin{axis}[
width=\columnwidth,
height=0.7\columnwidth,
xmin=0,
xmax=300,
ymin=-0.01,
ymax=.5,
xlabel={Training time $(t)$},
ylabel={Misclassification rate},
ytick={0,0.1,0.2,0.3,0.4,0.5},
grid=major,
scaled ticks=true,
]
\addplot[color=blue!60!white,line width=1pt] coordinates{
(0.000000,0.498444)(6.000000,0.121505)(12.000000,0.038776)(18.000000,0.017755)(24.000000,0.010153)(30.000000,0.006684)(36.000000,0.004821)(42.000000,0.003724)(48.000000,0.002602)(54.000000,0.002117)(60.000000,0.001760)(66.000000,0.001454)(72.000000,0.001046)(78.000000,0.000867)(84.000000,0.000714)(90.000000,0.000612)(96.000000,0.000485)(102.000000,0.000408)(108.000000,0.000332)(114.000000,0.000306)(120.000000,0.000281)(126.000000,0.000204)(132.000000,0.000179)(138.000000,0.000179)(144.000000,0.000179)(150.000000,0.000179)(156.000000,0.000179)(162.000000,0.000128)(168.000000,0.000051)(174.000000,0.000026)(180.000000,0.000026)(186.000000,0.000026)(192.000000,0.000026)(198.000000,0.000000)(204.000000,0.000000)(210.000000,0.000000)(216.000000,0.000000)(222.000000,0.000000)(228.000000,0.000000)(234.000000,0.000000)(240.000000,0.000000)(246.000000,0.000000)(252.000000,0.000000)(258.000000,0.000000)(264.000000,0.000000)(270.000000,0.000000)(276.000000,0.000000)(282.000000,0.000000)(288.000000,0.000000)(294.000000,0.000000)
};
\addlegendentry{{ Simulation: training performance }};
\addplot+[only marks, mark=x,color=blue!60!white] coordinates{
(0.000000,0.500000)(6.000000,0.126445)(12.000000,0.037929)(18.000000,0.016031)(24.000000,0.008300)(30.000000,0.004780)(36.000000,0.002917)(42.000000,0.001843)(48.000000,0.001192)(54.000000,0.000785)(60.000000,0.000524)(66.000000,0.000355)(72.000000,0.000244)(78.000000,0.000170)(84.000000,0.000119)(90.000000,0.000085)(96.000000,0.000061)(102.000000,0.000044)(108.000000,0.000032)(114.000000,0.000024)(120.000000,0.000018)(126.000000,0.000013)(132.000000,0.000010)(138.000000,0.000008)(144.000000,0.000006)(150.000000,0.000005)(156.000000,0.000004)(162.000000,0.000003)(168.000000,0.000002)(174.000000,0.000002)(180.000000,0.000001)(186.000000,0.000001)(192.000000,0.000001)(198.000000,0.000001)(204.000000,0.000001)(210.000000,0.000000)(216.000000,0.000000)(222.000000,0.000000)(228.000000,0.000000)(234.000000,0.000000)(240.000000,0.000000)(246.000000,0.000000)(252.000000,0.000000)(258.000000,0.000000)(264.000000,0.000000)(270.000000,0.000000)(276.000000,0.000000)(282.000000,0.000000)(288.000000,0.000000)(294.000000,0.000000)
};
\addlegendentry{{ Theory: training performance }};
\addplot[densely dashed,color=red!60!white,line width=1pt] coordinates{
(0.000000,0.502526)(6.000000,0.170995)(12.000000,0.080663)(18.000000,0.053954)(24.000000,0.043776)(30.000000,0.038546)(36.000000,0.035918)(42.000000,0.034184)(48.000000,0.033291)(54.000000,0.032730)(60.000000,0.032423)(66.000000,0.032551)(72.000000,0.032628)(78.000000,0.032883)(84.000000,0.033214)(90.000000,0.033367)(96.000000,0.033622)(102.000000,0.034082)(108.000000,0.034260)(114.000000,0.034668)(120.000000,0.035179)(126.000000,0.035689)(132.000000,0.036173)(138.000000,0.036786)(144.000000,0.037296)(150.000000,0.037857)(156.000000,0.038418)(162.000000,0.038776)(168.000000,0.039209)(174.000000,0.039668)(180.000000,0.040128)(186.000000,0.040561)(192.000000,0.040740)(198.000000,0.041148)(204.000000,0.041709)(210.000000,0.041964)(216.000000,0.042372)(222.000000,0.042781)(228.000000,0.043112)(234.000000,0.043469)(240.000000,0.044031)(246.000000,0.044490)(252.000000,0.045128)(258.000000,0.045536)(264.000000,0.045867)(270.000000,0.046199)(276.000000,0.046633)(282.000000,0.047092)(288.000000,0.047551)(294.000000,0.048061)
};
\addlegendentry{{ Simulation: generalization performance }};
\addplot+[only marks, mark=o,color=red!60!white] coordinates{
(0.000000,0.500000)(6.000000,0.174328)(12.000000,0.082045)(18.000000,0.053372)(24.000000,0.041642)(30.000000,0.035840)(36.000000,0.032617)(42.000000,0.030699)(48.000000,0.029518)(54.000000,0.028786)(60.000000,0.028345)(66.000000,0.028102)(72.000000,0.027999)(78.000000,0.027998)(84.000000,0.028075)(90.000000,0.028210)(96.000000,0.028392)(102.000000,0.028610)(108.000000,0.028858)(114.000000,0.029131)(120.000000,0.029423)(126.000000,0.029731)(132.000000,0.030053)(138.000000,0.030386)(144.000000,0.030728)(150.000000,0.031079)(156.000000,0.031436)(162.000000,0.031798)(168.000000,0.032164)(174.000000,0.032535)(180.000000,0.032908)(186.000000,0.033284)(192.000000,0.033661)(198.000000,0.034040)(204.000000,0.034420)(210.000000,0.034800)(216.000000,0.035181)(222.000000,0.035562)(228.000000,0.035943)(234.000000,0.036324)(240.000000,0.036704)(246.000000,0.037084)(252.000000,0.037463)(258.000000,0.037840)(264.000000,0.038217)(270.000000,0.038593)(276.000000,0.038968)(282.000000,0.039341)(288.000000,0.039713)(294.000000,0.040084)
};
\addlegendentry{{ Theory: generalization performance }};
\begin{scope}
    \spy[black!50!white,size=1.6cm,circle,connect spies,magnification=2] on (0.6,0.4) in node [fill=none] at (4,1.5);
\end{scope}
\end{axis}
\end{tikzpicture}
\caption{Training and generalization performance for MNIST data (number $1$ and $7$) with $n=p=784$, $c_1 = c_2 = 1/2$, $\alpha=0.01$ and $\sigma^2 = 0.1$. Results obtained by averaging over $100$ runs.}
\label{fig:MNIST-simu}
\end{center}
\vskip -0.1in
\end{figure}
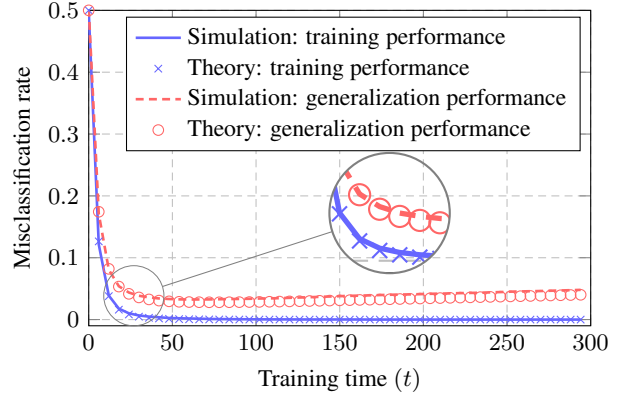

\section{Numerical Validations}
\label{sec:validations}

We close this article with experiments on the popular MNIST dataset \cite{lecun1998mnist} (number $1$ and $7$). We randomly select training sets of size $n=784$ vectorized images of dimension $p=784$ and add artificially a Gaussian white noise of $-10\si{\deci\bel}$ in order to be more compliant with our toy model setting. Empirical means and covariances of each class are estimated from the full set of $13\,007$ MNIST images ($6\,742$ images of number $1$ and $6\,265$ of number $7$). The image vectors in each class are whitened by pre-multiplying $\C_a^{-1/2}$ and re-centered to have means of $\pm \bmu$, with $\bmu$ half of the difference between means from the two classes. We observe an extremely close fit between our results and the empirical simulations, as shown in Figure~\ref{fig:MNIST-simu}.

\section{Conclusion}
\label{sec:conclusion}

In this article, we established a random matrix approach to the analysis of learning dynamics for gradient-based algorithms on data of simultaneously large dimension and size. With a toy model of Gaussian mixture data with $\pm \bmu$ means and identity covariance, we have shown that the training and generalization performances of the network have asymptotically deterministic behaviors that can be evaluated via so-called deterministic equivalents and computed with complex contour integrals (and even under the form of real integrals in the present setting). The article can be generalized in many ways: with more generic mixture models (with the Gaussian assumption relaxed), on more appropriate loss functions (logistic regression for example), and more advanced optimization methods.
%This result can be taken as a first step into the analysis of more elaborate network structures or data model

In the present work, the analysis has been performed on the ``full-batch'' gradient descent system. However, the most popular method used today is in fact its ``stochastic'' version \cite{bottou2010large} where only a fixed-size ($n_{batch}$) randomly selected subset (called a \emph{mini-batch}) of the training data is used to compute the gradient and descend \emph{one} step along with the opposite direction of this gradient in each iteration. In this scenario, one of major concern in practice lies in determining the optimal size of the mini-batch and its influence on the generalization performance of the network \cite{keskar2016large}. This can be naturally linked to the ratio $n_{batch}/p$ in the random matrix analysis.

Deep networks that are of more practical interests, however, need more efforts. As mentioned in \cite{saxe2013exact,advani2017high}, in the case of multi-layer networks, the learning dynamics depend, instead of each eigenmode separately, on the coupling of different eigenmodes from different layers. To handle this difficulty, one may add extra assumptions of independence between layers as in \cite{choromanska2015loss} so as to study each layer separately and then reassemble to retrieve the results of the whole network.

%$mini-batch setting, multi-layer with assumptions of the independence between layers, simple nonlinearity 

% \subsection{Extension to more general data model}

% Using the results of deterministic equivalents found in \cite{benaych2016spectral} for $K$-classes Gaussian mixture model of $\mathcal{N}(\bmu_a, \mathbf{C}_a)$, with $a = 1,\ldots,K$, one is able to generalize current results to GMM data that are of simultaneously large dimension and size. This study may shed new light on the contribution from covariance structure of the data, since in current study we assume the covariance to be identity, and only the influence of means is studied.

% \subsection{Extension to nonlinear activation}

% One of the main challenges of applying random matrix results to modern deep neural nets lies on the non-linearities of activation functions. However, in recent line of works \cite{el2010spectrum,cheng2013spectrum,louart2017random,pennington2017nonlinear} several methods have been considered to handle this problem of non-linearity in the contexts of kernel methods, random feature maps or neural nets. Taking advantages of these results one may extend current results to a non-linear setting and have a deeper understanding on the choice of non-linear activation.
\section*{Acknowledgments}

We thank the anonymous reviewers for their comments and constructive suggestions. We would like to acknowledge this work is supported by the ANR Project RMT4GRAPH (ANR-14-CE28-0006) and the Project DeepRMT of La Fondation Sup{\'e}lec.

\bibliography{liao}
\bibliographystyle{icml2018}

\clearpage

\appendix
\onecolumn

\begin{center}
  {\Large \textbf{Supplementary Material\\}} \vskip 0.1in \textbf{The Dynamics of Learning: A Random Matrix Approach}
\end{center}
\vskip 0.3in

\section{Proofs}
\label{sm:detailed-deduction}

\subsection{Proofs of Theorem~\ref{theo:generalize-perf}~and~\ref{theo:training-perf}}
\label{sm:proofs-of-theo}

\begin{proof}
We start with the proof of Theorem~\ref{theo:generalize-perf}, since
\begin{align*}
	&\bmu^\T \w(t) = \bmu^\T e^{- \frac{\alpha t}n \X \X^\T } \w_0 + \bmu^\T \left(\I_p - e^{- \frac{\alpha t}n \X\X^\T } \right) \w_{LS}\\
	%%%
	&= -\frac1{2\pi i} \oint_{\gamma} f_t(z) \bmu^\T \left( \frac1n \X \X^\T - z \I_p \right)^{-1} \w_0 \ dz - \frac1{2\pi i} \oint_{\gamma} \frac{1-f_t(z)}{z} \bmu^\T \left( \frac1n \X \X^\T - z \I_p \right)^{-1} \frac1n \X\y \ dz
\end{align*}
with $ \frac1n \X \X^\T = \frac1n \Z \Z^\T + \begin{bmatrix} \bmu & \frac1n \Z\y \end{bmatrix} \begin{bmatrix} 1 & 1 \\ 1 & 0 \end{bmatrix} \begin{bmatrix} \bmu^\T \\ \frac1n \y^\T \Z^\T \end{bmatrix}$ and therefore
\[
	\left( \frac1n \X \X^\T - z \I_p \right)^{-1} = \Q(z) - \Q(z) \begin{bmatrix} \bmu & \frac1n \Z\y \end{bmatrix} \begin{bmatrix} \bmu^\T \Q(z) \bmu & 1+\frac1n \bmu^\T \Q(z) \Z\y \\ 1+\frac1n \bmu^\T \Q(z) \Z\y & -1 + \frac1n \y^\T \Z^\T \Q(z) \frac1n \Z \y \end{bmatrix}^{-1} \begin{bmatrix} \bmu^\T \\ \frac1n \y^\T \Z^\T \end{bmatrix} \Q(z).
\]

We thus resort to the computation of the bilinear form $\mathbf{a}^\T \Q(z) \mathbf{b}$, for which we plug-in the deterministic equivalent of $\Q(z) \leftrightarrow \bar \Q(z) = m(z) \I_p$ to obtain the following estimations
\begin{align*}
	\bmu^\T \Q(z) \bmu & = \| \bmu \|^2 m(z) \\
	%%%
	\frac1n \bmu^\T \Q(z) \Z\y & = o(1) \\
	%%%
	\frac1{n^2} \y^\T \Z^\T \Q(z) \Z\y & = \frac1{n^2} \y^\T \tilde \Q(z) \Z^\T \Z \y = \frac1n \y^\T \tilde \Q(z) \left( \frac1n \Z^\T \Z - z \I_n + z \I_n \right) \y \\
	%%%
	&= \frac1n \| \y\|^2 + z \frac1n \y^\T \tilde \Q(z) \y = 1 + z \frac1n \tr \tilde \Q(z) = 1 + z \tilde m(z)
\end{align*}
with the \emph{co-resolvent} $\tilde \Q(z) = \left( \frac1n \Z^\T \Z - z \I_n \right)^{-1}$, $m(z)$ the \emph{unique} solution of the Marčenko–Pastur equation \eqref{eq:MP-equation} and $\tilde m(z) = \frac1n \tr \tilde \Q(z) + o(1)$ such that
\[
	c m(z) = \tilde m(z) + \frac1z (1-c) 
\]
which is a direct result of the fact that both $\Z^\T \Z$ and $\Z \Z^\T$ have the same eigenvalues except for the additional zeros eigenvalues for the larger matrix (which essentially depends on the sign of $1-c$).

We thus get, with the Schur complement lemma,
\begin{align*}
	\left( \frac1n \X \X^\T - z \I_p \right)^{-1} &= \Q(z) - \Q(z) \begin{bmatrix} \bmu & \frac1n \Z\y \end{bmatrix} \begin{bmatrix} \|\bmu\|^2 m(z) & 1 \\ 1 & z \tilde m(z) \end{bmatrix}^{-1} \begin{bmatrix} \bmu^\T \\ \frac1n \y^\T \Z^\T \end{bmatrix} \Q(z) + o(1) \\
	%%%
	& = \Q(z) - \frac{ \Q(z) }{ z \|\bmu\|^2 m(z) \tilde m(z) - 1 } \begin{bmatrix} \bmu & \frac1n \Z\y \end{bmatrix} \begin{bmatrix} z \tilde m(z) & -1 \\ -1 & \|\bmu\|^2 m(z) \end{bmatrix} \begin{bmatrix} \bmu^\T \\ \frac1n \y^\T \Z^\T \end{bmatrix} \Q(z) + o(1)
\end{align*}
and the term $\bmu^\T \left( \frac1n \X \X^\T - z \I_p \right)^{-1} \frac1n \X\y$ is therefore given by
\begin{align*}
	&\bmu^\T \left( \frac1n \X \X^\T - z \I_p \right)^{-1} \frac1n \X\y = \| \bmu \|^2 m(z) - \frac{ \begin{bmatrix} \| \bmu \|^2 m(z) & 0 \end{bmatrix} }{ z \| \bmu \|^2 m(z) \tilde m(z) -1 }  \begin{bmatrix} z \tilde m(z) & -1 \\ -1 & \| \bmu \|^2 m(z) \end{bmatrix} \begin{bmatrix} \| \bmu \|^2 m(z) \\ 1 + z \tilde m(z) \end{bmatrix} + o(1)\\
	&= \frac{\| \bmu \|^2 m(z) z \tilde m(z) }{ \| \bmu \|^2 m(z) z \tilde m(z) -1 } + o(1) = \frac{\| \bmu \|^2 (z m(z) + 1)}{ 1 + \| \bmu \|^2 (z m(z) + 1) } + o(1) = \frac{ \| \bmu\|^2 m(z) }{ (\| \bmu\|^2 +c ) m(z) +1 } + o(1)
\end{align*}
where we use the fact that $\tilde m(z) = c m(z) - \frac1z (1-c) $ and $ (zm(z) + 1) (c m(z) + 1) = m$ from \eqref{eq:MP-equation}, while the term $\bmu^\T \left( \frac1n \X \X^\T - z \I_p \right)^{-1} \w_0 = O(n^{-\frac12})$ due to the independence of $\w_0$ with respect to $\Z$ and can be check with a careful application of Lyapunov’s central limit theorem \cite{billingsley2008probability}.

Following the same arguments we have
\begin{align*}
	& \w(t)^\T \w(t) = -\frac1{2\pi i} \oint_{\gamma} f^2_t(z) \w_0 \left( \frac1n \X\X^\T - z \I_p \right)^{-1} \w_0 \ dz -\frac1{\pi i} \oint_{\gamma} \frac{f_t(z) (1- f_t(z))}{z} \w_0 \left( \frac1n \X\X^\T - z \I_p \right)^{-1} \frac1n \X \y \ dz \\
	%%%
	&- \frac1{2\pi i} \oint_{\gamma} \frac{ (1- f_t(z))^2 }{z^2} \frac1n \y^\T \X^\T \left( \frac1n \X\X^\T - z \I_p \right)^{-1} \frac1n \X \y \ dz
\end{align*}
together with
\begin{align*}
	\w_0 \left( \frac1n \X\X^\T - z \I_p \right)^{-1} \w_0 & = \sigma^2 m(z) + o(1) \\
	%%%
	\w_0 \left( \frac1n \X\X^\T - z \I_p \right)^{-1} \frac1n \X \y^\T &= o(1) \\
	%%%
	\frac1n \y^\T \X^\T \left( \frac1n \X\X^\T - z \I_p \right)^{-1} \frac1n \X \y &= 1 - \frac1{ (\| \bmu\|^2 +c) m(z) +1  } + o(1).
\end{align*}

It now remains to replace the different terms in $\bmu^\T \w(t)$ and $\w(t)^\T \w(t)$ by their asymptotic approximations. To this end, first note that all aforementioned approximations can be summarized as the fact that, for a generic $h(z)$, we have, as $n \to \infty$,
\[
	h(z) - \bar h(z) \to 0
\]
almost surely for all $z$ not an eigenvalue of $\frac1n \X\X^\T$. Therefore, there exists a probability one set $\Omega_z$ on which $h(z)$ is uniformly bounded for all large $n$, with a bound independent of $z$. Then by the Theorem of ``no eigenvalues outside the support'' (see for example \cite{bai1998no}) we know that, with probability one, for all $n,p$ large, no eigenvalue of $\frac1n \Z\Z^\T$ appears outside the interval $[\lambda_-,\lambda_+]$, where we recall $\lambda_- \equiv (1-\sqrt{c})^2$ and $\lambda_+ \equiv (1+\sqrt{c})^2$. As such, the set of intersection $\Omega = \cap_{z_i} \Omega_{z_i}$ for a finitely many $z_i$, is still a probability one set. Finally by Vitali convergence theorem, together with the analyticity of the function under consideration, we conclude the proof of Theorem~\ref{theo:generalize-perf}. The proof of Theorem~\ref{theo:training-perf} follows exactly the same line of arguments and is thus omitted here.
\end{proof}

\subsection{Detailed Derivation of \eqref{eq:E}-\eqref{eq:Var-star}}
\label{sm:detailed-computation-E-V}

%%% eigenvalue distribution and MP law
\begin{figure}[htb]
\vskip 0.1in
\begin{center}
\begin{tikzpicture}[font=\footnotesize]
\begin{axis}[
	width=0.5\columnwidth,
	height=0.35\columnwidth,
    xmin=-.5,
    xmax=5,
    ymin=-0.5,
    ymax=0.5,
    axis equal,
    axis lines=middle,
    xlabel=$\Re(z)$,
    ylabel=$\Im(z)$,
    disabledatascaling]
    \addplot+[only marks,mark =x,color=blue!60!white] coordinates{
    (0.092548, 0.000000)(0.095899, 0.000000)(0.098369, 0.000000)(0.102544, 0.000000)(0.105409, 0.000000)(0.106708, 0.000000)(0.110633, 0.000000)(0.111364, 0.000000)(0.113748, 0.000000)(0.118289, 0.000000)(0.118810, 0.000000)(0.121582, 0.000000)(0.123659, 0.000000)(0.126776, 0.000000)(0.129911, 0.000000)(0.130256, 0.000000)(0.132252, 0.000000)(0.136355, 0.000000)(0.138399, 0.000000)(0.141040, 0.000000)(0.141719, 0.000000)(0.144760, 0.000000)(0.147218, 0.000000)(0.147668, 0.000000)(0.149468, 0.000000)(0.150711, 0.000000)(0.153436, 0.000000)(0.155468, 0.000000)(0.158899, 0.000000)(0.161894, 0.000000)(0.162799, 0.000000)(0.163379, 0.000000)(0.166721, 0.000000)(0.167652, 0.000000)(0.171421, 0.000000)(0.173714, 0.000000)(0.175160, 0.000000)(0.177191, 0.000000)(0.179805, 0.000000)(0.182530, 0.000000)(0.185374, 0.000000)(0.185845, 0.000000)(0.189680, 0.000000)(0.192354, 0.000000)(0.193496, 0.000000)(0.196592, 0.000000)(0.197761, 0.000000)(0.201265, 0.000000)(0.203395, 0.000000)(0.206171, 0.000000)(0.207612, 0.000000)(0.209977, 0.000000)(0.211941, 0.000000)(0.214253, 0.000000)(0.214611, 0.000000)(0.219491, 0.000000)(0.221185, 0.000000)(0.222785, 0.000000)(0.225077, 0.000000)(0.227927, 0.000000)(0.229576, 0.000000)(0.232260, 0.000000)(0.233340, 0.000000)(0.235048, 0.000000)(0.236344, 0.000000)(0.239496, 0.000000)(0.242720, 0.000000)(0.246679, 0.000000)(0.248871, 0.000000)(0.251803, 0.000000)(0.252941, 0.000000)(0.253637, 0.000000)(0.256828, 0.000000)(0.259942, 0.000000)(0.261193, 0.000000)(0.262697, 0.000000)(0.266118, 0.000000)(0.267641, 0.000000)(0.269025, 0.000000)(0.271258, 0.000000)(0.275644, 0.000000)(0.277972, 0.000000)(0.279059, 0.000000)(0.283268, 0.000000)(0.285556, 0.000000)(0.288041, 0.000000)(0.290317, 0.000000)(0.290805, 0.000000)(0.296129, 0.000000)(0.296793, 0.000000)(0.301580, 0.000000)(0.302344, 0.000000)(0.303169, 0.000000)(0.305986, 0.000000)(0.308491, 0.000000)(0.311061, 0.000000)(0.313376, 0.000000)(0.318368, 0.000000)(0.321105, 0.000000)(0.324588, 0.000000)(0.325193, 0.000000)(0.326761, 0.000000)(0.329992, 0.000000)(0.331715, 0.000000)(0.333597, 0.000000)(0.339480, 0.000000)(0.340389, 0.000000)(0.344558, 0.000000)(0.346651, 0.000000)(0.348068, 0.000000)(0.350412, 0.000000)(0.352292, 0.000000)(0.356334, 0.000000)(0.359434, 0.000000)(0.363438, 0.000000)(0.366894, 0.000000)(0.372782, 0.000000)(0.373531, 0.000000)(0.375849, 0.000000)(0.378914, 0.000000)(0.381273, 0.000000)(0.382568, 0.000000)(0.384062, 0.000000)(0.386772, 0.000000)(0.388477, 0.000000)(0.393226, 0.000000)(0.394623, 0.000000)(0.397183, 0.000000)(0.398857, 0.000000)(0.401436, 0.000000)(0.404643, 0.000000)(0.406613, 0.000000)(0.407013, 0.000000)(0.412564, 0.000000)(0.415458, 0.000000)(0.417477, 0.000000)(0.422351, 0.000000)(0.426884, 0.000000)(0.430196, 0.000000)(0.432570, 0.000000)(0.436577, 0.000000)(0.437742, 0.000000)(0.442343, 0.000000)(0.444940, 0.000000)(0.448103, 0.000000)(0.451424, 0.000000)(0.454273, 0.000000)(0.454957, 0.000000)(0.457902, 0.000000)(0.460155, 0.000000)(0.462252, 0.000000)(0.466638, 0.000000)(0.470583, 0.000000)(0.472561, 0.000000)(0.475431, 0.000000)(0.477207, 0.000000)(0.481336, 0.000000)(0.482923, 0.000000)(0.485170, 0.000000)(0.486563, 0.000000)(0.489881, 0.000000)(0.491161, 0.000000)(0.496642, 0.000000)(0.498329, 0.000000)(0.506177, 0.000000)(0.508311, 0.000000)(0.509861, 0.000000)(0.511208, 0.000000)(0.518939, 0.000000)(0.521272, 0.000000)(0.522788, 0.000000)(0.525763, 0.000000)(0.527882, 0.000000)(0.532040, 0.000000)(0.534476, 0.000000)(0.535349, 0.000000)(0.541288, 0.000000)(0.544313, 0.000000)(0.547820, 0.000000)(0.553367, 0.000000)(0.557107, 0.000000)(0.559476, 0.000000)(0.563069, 0.000000)(0.563858, 0.000000)(0.567703, 0.000000)(0.571372, 0.000000)(0.574529, 0.000000)(0.578207, 0.000000)(0.579017, 0.000000)(0.583118, 0.000000)(0.587216, 0.000000)(0.591672, 0.000000)(0.592543, 0.000000)(0.595504, 0.000000)(0.602581, 0.000000)(0.604685, 0.000000)(0.606219, 0.000000)(0.611824, 0.000000)(0.612141, 0.000000)(0.617783, 0.000000)(0.619309, 0.000000)(0.621576, 0.000000)(0.625899, 0.000000)(0.633575, 0.000000)(0.635536, 0.000000)(0.639505, 0.000000)(0.641581, 0.000000)(0.643311, 0.000000)(0.648218, 0.000000)(0.652146, 0.000000)(0.655581, 0.000000)(0.659199, 0.000000)(0.662990, 0.000000)(0.666947, 0.000000)(0.670386, 0.000000)(0.677949, 0.000000)(0.680021, 0.000000)(0.681404, 0.000000)(0.682039, 0.000000)(0.686198, 0.000000)(0.691369, 0.000000)(0.693378, 0.000000)(0.698281, 0.000000)(0.700853, 0.000000)(0.710864, 0.000000)(0.712966, 0.000000)(0.717339, 0.000000)(0.719366, 0.000000)(0.724363, 0.000000)(0.726531, 0.000000)(0.727224, 0.000000)(0.732757, 0.000000)(0.737324, 0.000000)(0.739844, 0.000000)(0.746061, 0.000000)(0.752453, 0.000000)(0.755243, 0.000000)(0.758085, 0.000000)(0.760912, 0.000000)(0.762166, 0.000000)(0.769171, 0.000000)(0.773949, 0.000000)(0.779007, 0.000000)(0.784316, 0.000000)(0.785582, 0.000000)(0.788104, 0.000000)(0.793284, 0.000000)(0.794451, 0.000000)(0.797365, 0.000000)(0.800827, 0.000000)(0.801767, 0.000000)(0.808343, 0.000000)(0.812694, 0.000000)(0.814293, 0.000000)(0.819289, 0.000000)(0.822054, 0.000000)(0.830725, 0.000000)(0.837210, 0.000000)(0.842441, 0.000000)(0.843986, 0.000000)(0.848032, 0.000000)(0.852750, 0.000000)(0.859587, 0.000000)(0.862844, 0.000000)(0.864180, 0.000000)(0.870897, 0.000000)(0.874771, 0.000000)(0.877396, 0.000000)(0.882393, 0.000000)(0.886394, 0.000000)(0.888558, 0.000000)(0.897980, 0.000000)(0.900564, 0.000000)(0.904167, 0.000000)(0.912000, 0.000000)(0.912917, 0.000000)(0.922172, 0.000000)(0.923551, 0.000000)(0.926584, 0.000000)(0.927689, 0.000000)(0.931632, 0.000000)(0.935737, 0.000000)(0.941329, 0.000000)(0.945425, 0.000000)(0.947678, 0.000000)(0.950019, 0.000000)(0.957095, 0.000000)(0.960566, 0.000000)(0.965456, 0.000000)(0.969920, 0.000000)(0.976557, 0.000000)(0.981665, 0.000000)(0.986018, 0.000000)(0.991523, 0.000000)(0.994493, 0.000000)(1.003692, 0.000000)(1.006194, 0.000000)(1.008617, 0.000000)(1.014723, 0.000000)(1.020056, 0.000000)(1.022783, 0.000000)(1.025178, 0.000000)(1.028900, 0.000000)(1.038638, 0.000000)(1.039680, 0.000000)(1.049708, 0.000000)(1.050424, 0.000000)(1.052806, 0.000000)(1.057754, 0.000000)(1.062325, 0.000000)(1.069861, 0.000000)(1.075437, 0.000000)(1.078313, 0.000000)(1.086915, 0.000000)(1.091537, 0.000000)(1.098136, 0.000000)(1.102359, 0.000000)(1.107828, 0.000000)(1.112299, 0.000000)(1.114543, 0.000000)(1.121459, 0.000000)(1.129235, 0.000000)(1.132790, 0.000000)(1.141850, 0.000000)(1.147625, 0.000000)(1.148563, 0.000000)(1.152700, 0.000000)(1.154085, 0.000000)(1.162026, 0.000000)(1.166794, 0.000000)(1.172130, 0.000000)(1.180688, 0.000000)(1.186466, 0.000000)(1.194891, 0.000000)(1.197584, 0.000000)(1.201585, 0.000000)(1.207570, 0.000000)(1.209453, 0.000000)(1.219015, 0.000000)(1.225800, 0.000000)(1.227081, 0.000000)(1.232816, 0.000000)(1.235490, 0.000000)(1.241327, 0.000000)(1.243009, 0.000000)(1.248624, 0.000000)(1.261858, 0.000000)(1.266029, 0.000000)(1.269454, 0.000000)(1.274206, 0.000000)(1.280179, 0.000000)(1.281160, 0.000000)(1.290290, 0.000000)(1.292633, 0.000000)(1.299733, 0.000000)(1.303985, 0.000000)(1.312151, 0.000000)(1.321103, 0.000000)(1.328836, 0.000000)(1.335451, 0.000000)(1.341365, 0.000000)(1.343390, 0.000000)(1.353892, 0.000000)(1.356013, 0.000000)(1.366915, 0.000000)(1.371492, 0.000000)(1.375442, 0.000000)(1.379768, 0.000000)(1.382928, 0.000000)(1.391654, 0.000000)(1.393210, 0.000000)(1.407431, 0.000000)(1.412884, 0.000000)(1.417258, 0.000000)(1.425763, 0.000000)(1.429789, 0.000000)(1.432332, 0.000000)(1.442231, 0.000000)(1.446941, 0.000000)(1.452584, 0.000000)(1.455576, 0.000000)(1.461764, 0.000000)(1.469448, 0.000000)(1.480817, 0.000000)(1.485334, 0.000000)(1.498991, 0.000000)(1.505256, 0.000000)(1.513538, 0.000000)(1.515564, 0.000000)(1.516590, 0.000000)(1.525994, 0.000000)(1.537118, 0.000000)(1.537899, 0.000000)(1.546089, 0.000000)(1.549202, 0.000000)(1.558577, 0.000000)(1.560170, 0.000000)(1.563789, 0.000000)(1.578343, 0.000000)(1.592569, 0.000000)(1.593845, 0.000000)(1.609331, 0.000000)(1.615200, 0.000000)(1.622401, 0.000000)(1.626033, 0.000000)(1.630808, 0.000000)(1.637880, 0.000000)(1.647533, 0.000000)(1.656309, 0.000000)(1.659897, 0.000000)(1.677448, 0.000000)(1.684582, 0.000000)(1.690100, 0.000000)(1.693967, 0.000000)(1.697205, 0.000000)(1.713276, 0.000000)(1.720048, 0.000000)(1.728945, 0.000000)(1.734687, 0.000000)(1.739075, 0.000000)(1.743133, 0.000000)(1.756304, 0.000000)(1.760726, 0.000000)(1.775198, 0.000000)(1.778566, 0.000000)(1.780390, 0.000000)(1.790197, 0.000000)(1.794137, 0.000000)(1.800465, 0.000000)(1.814337, 0.000000)(1.820609, 0.000000)(1.826629, 0.000000)(1.839217, 0.000000)(1.845300, 0.000000)(1.854449, 0.000000)(1.865117, 0.000000)(1.872521, 0.000000)(1.885124, 0.000000)(1.886608, 0.000000)(1.893448, 0.000000)(1.904904, 0.000000)(1.916330, 0.000000)(1.926176, 0.000000)(1.933920, 0.000000)(1.943936, 0.000000)(1.954993, 0.000000)(1.962927, 0.000000)(1.972880, 0.000000)(1.976573, 0.000000)(1.986917, 0.000000)(1.999548, 0.000000)(2.002835, 0.000000)(2.017696, 0.000000)(2.024490, 0.000000)(2.036751, 0.000000)(2.038062, 0.000000)(2.046032, 0.000000)(2.058542, 0.000000)(2.070398, 0.000000)(2.075096, 0.000000)(2.084516, 0.000000)(2.090540, 0.000000)(2.104909, 0.000000)(2.117206, 0.000000)(2.126271, 0.000000)(2.141972, 0.000000)(2.151987, 0.000000)(2.157567, 0.000000)(2.167949, 0.000000)(2.180084, 0.000000)(2.182474, 0.000000)(2.200334, 0.000000)(2.212876, 0.000000)(2.216601, 0.000000)(2.222157, 0.000000)(2.249534, 0.000000)(2.260769, 0.000000)(2.267221, 0.000000)(2.273791, 0.000000)(2.283293, 0.000000)(2.299963, 0.000000)(2.315438, 0.000000)(2.327745, 0.000000)(2.331705, 0.000000)(2.348536, 0.000000)(2.364218, 0.000000)(2.368660, 0.000000)(2.398058, 0.000000)(2.407767, 0.000000)(2.428139, 0.000000)(2.430867, 0.000000)(2.446505, 0.000000)(2.457356, 0.000000)(2.470891, 0.000000)(2.489893, 0.000000)(2.503496, 0.000000)(2.525401, 0.000000)(2.553036, 0.000000)(2.554570, 0.000000)(2.566282, 0.000000)(2.572356, 0.000000)(2.596883, 0.000000)(2.611552, 0.000000)(2.630440, 0.000000)(2.665859, 0.000000)(2.693072, 0.000000)(2.715374, 0.000000)(2.731692, 0.000000)(2.749714, 0.000000)(2.765227, 0.000000)(2.799732, 0.000000)(2.881722, 0.000000)(3.9722,0)
	};
	\addlegendentry{{Eigenvalues of $\frac1n \X\X^\T$}};
	%%%contour
	\draw [->,color=red!60!white,line width=1pt] (-0.1,-0.5) -- (1.5,-0.5) node [below] {$\gamma_b$};
	\draw [color=red!60!white,line width=1pt] (1.5,-0.5) -- (3.2,-0.5);
	%%%
	\draw [->,color=red!60!white,line width=1pt] (3.2,-0.5) -- (3.2,0);
	\draw [color=red!60!white,line width=1pt] (3.2,0) -- (3.2,0.5);
	%%%
	\draw [->,color=red!60!white,line width=1pt] (3.2,0.5) -- (1.5,0.5);
	\draw [color=red!60!white,line width=1pt] (1.5,0.5) -- (-0.1,0.5);
	%%%
	\draw [->,color=red!60!white,line width=1pt] (-0.1,0.5) -- (-0.1,0);
	\draw [color=red!60!white,line width=1pt] (-0.1,0) -- (-0.1,-0.5);
	%
	%
	%
	%
	%
	%%%
	\draw [->,color=red!60!white,line width=1pt] (3.25,-0.5) -- (4.2,-0.5) node [below] {$\gamma_s$};
	%%%
	\draw [color=red!60!white,line width=1pt] (3.25,-0.5) -- (3.25,0);
	\draw [<-,color=red!60!white,line width=1pt] (3.25,0) -- (3.25,0.5);
	%%%
	\draw [->,color=red!60!white,line width=1pt] (4.2,-0.5) -- (4.2,0);
	\draw [color=red!60!white,line width=1pt] (4.2,0) -- (4.2,0.5);
	%%%
	\draw [->,color=red!60!white,line width=1pt] (4.2,0.5) -- (3.25,0.5);
	\addlegendimage{->,line width=1pt,color=red!60!white}
	\addlegendentry{{Integration path $\gamma$}};
	%%%
	\draw [<->,line width=1pt] (-0.2,0.5) node [left,yshift=-0.02\columnwidth] { $\varepsilon$ } -- (-0.2,0);
	\draw [<->,line width=1pt] (-0.2,-0.1) node [below,xshift=0.02\columnwidth] { $\epsilon$ } -- (0.1,-0.1);
	\draw [<->,line width=1pt] (2.9,-0.1) node [below,xshift=0.02\columnwidth] { $\epsilon$ } -- (3.2,-0.1);
\end{axis}
\end{tikzpicture}
\caption{Eigenvalue distribution of $\frac1n \X\X^\T$ for $\bmu = [1.5;\mathbf{0}_{p-1}]$, $p=512$, $n=1\,024$ and $c_1 = c_2 = 1/2$.}
\label{sm:fig:eigvenvalue-distribution-XX}
\end{center}
\vskip -0.1in
\end{figure}
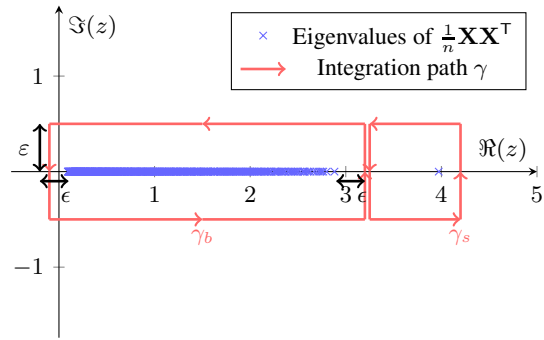

We first determine the location of the isolated eigenvalue $\lambda$ (as shown in Figure~\ref{fig:eigvenvalue-distribution-XX}). More concretely, we would like to find $\lambda$ an eigenvalue of $\frac1n \X\X^\T$ that lies outside the support of Marčenko–Pastur distribution (in fact, not an eigenvalue of $\frac1n \Z\Z^\T$). Solving the following equation for $\lambda \in \mathbb{R}$,
\begin{align*}
	&\det \left( \frac1n \X\X^\T - \lambda \I_p \right) =0 \\
	%%%
	&\Leftrightarrow \det\left( \frac1n \Z\Z^\T - \lambda \I_p + \begin{bmatrix} \bmu & \frac1n \Z\y \end{bmatrix} \begin{bmatrix} 1 & 1 \\ 1 & 0 \end{bmatrix} \begin{bmatrix} \bmu^\T \\ \frac1n \y^\T \Z^\T \end{bmatrix} \right) = 0 \\
	%%%
	&\Leftrightarrow \det \left( \frac1n \Z\Z^\T - \lambda \I_p \right) \det \left( \I_p + \Q(\lambda) \begin{bmatrix} \bmu & \frac1n \Z\y \end{bmatrix} \begin{bmatrix} 1 & 1 \\ 1 & 0 \end{bmatrix} \begin{bmatrix} \bmu^\T \\ \frac1n \y^\T \Z^\T \end{bmatrix} \right) = 0 \\
	%%%
	&\Leftrightarrow \det \left( \I_2 + \begin{bmatrix} 1 & 1 \\ 1 & 0 \end{bmatrix} \begin{bmatrix} \bmu^\T \\ \frac1n \y^\T \Z^\T \end{bmatrix} \Q(\lambda) \begin{bmatrix} \bmu & \frac1n \Z\y \end{bmatrix} \right) = 0 \\
	%%%
	&\Leftrightarrow \det \begin{bmatrix} \|\bmu\|^2 m(\lambda) + 1 & 1 + z \tilde m(\lambda) \\ \|\bmu\|^2 m(\lambda) & 1\end{bmatrix} + o(1) = 0 \\
	%%%
	&\Leftrightarrow 1 + ( \| \bmu\|^2 + c ) m(\lambda) + o(1) = 0
\end{align*}
where we recall the definition $\Q(\lambda) \equiv \left( \frac1n \Z\Z^\T - \lambda \I_p \right)^{-1}$ and use the fact that $\det(\mathbf{A} \mathbf{B}) = \det(\mathbf{A}) \det(\mathbf{B})$ as well as the Sylvester's determinant identity $\det(\I_p + \mathbf{A} \mathbf{B}) = \det( \I_n + \mathbf{B} \mathbf{A})$ for $\mathbf{A,B}$ of appropriate dimension. Together with \eqref{eq:MP-equation} we deduce the (empirical) isolated eigenvalue $\lambda = \lambda_s + o(1)$ with
\[
	\lambda_s = c+1 + \| \bmu\|^2 + \frac{c}{ \| \bmu\|^2 }
\]
which in fact gives the asymptotic location of the isolated eigenvalue as $n \to \infty$. In the following, we may thus use $\lambda_s$ instead of $\lambda$ throughout the computation. By splitting the path $\gamma$ into $\gamma_b+\gamma_s$ that circles respectively around the main bulk between $[\lambda_-,\lambda_+]$ and the isolated eigenvalue $\lambda_s$, we easily deduce, with the residual theorem that $E = E_{\gamma_b} + E_{\gamma_s}$ with
\begin{align}
	E_{\gamma_s} &= -\frac1{2\pi i} \oint_{\gamma_s} \frac{1-f_t(z)}z \frac{ \| \bmu \|^2 m(z)}{ 1 + (\| \bmu \|^2 + c) m(z) }\ dz = - {\rm Res} \frac{1-f_t(z)}z \frac{ \| \bmu \|^2 m(z)}{ 1 + (\| \bmu \|^2 + c) m (z) } \nonumber \\
	&= - \lim_{z \to \lambda_s} (z - \lambda_s) \frac{1-f_t(z)}z \frac{ \| \bmu \|^2 m(z)}{ 1 + (\| \bmu \|^2 + c) m (z) } = - \frac{1-f_t(\lambda_s)}{\lambda_s} \frac{ \| \bmu \|^2 m(\lambda_s)}{ (\| \bmu \|^2 + c) m^\prime(\lambda_s) } \nonumber \\
	%%%
	&= -\frac{\| \bmu\|^2}{\| \bmu\|^2 + c} \frac{1-f_t(\lambda_s)}{\lambda_s} \frac{1 - c - \lambda_s - 2c\lambda_s m(\lambda_s)}{c m(\lambda_s) +1} = \left( \| \bmu \|^2 - \frac{c}{\| \bmu \|^2} \right) \frac{1-f_t(\lambda_s)}{\lambda_s} \label{eq:E-gamma-s}
\end{align}
with $m^\prime(z)$ the derivative of $m(z)$ with respect to $z$ and is obtained by taking the derivative of \eqref{eq:MP-equation}.

We now move on to handle the contour integration $\gamma_b$ in the computation of $E_{\gamma_b}$. We follow the idea in \cite{bai2008clt} and choose the contour $\gamma_b$ to be a rectangle with sides parallel to the axes, intersecting the real axis at $0$ and $\lambda_+$ (in fact at $-\epsilon$ and $\lambda_+ + \epsilon$ so that the functions under consideration remain analytic) and the horizontal sides being a distance $\varepsilon \to 0$ away from the real axis. Since for nonzero $x \in \mathbb{R}$, the limit $\lim_{z \in \mathbb{Z} \to x} m(z) \equiv \check m(x)$ exists \cite{silverstein1995analysis} and is given by
\[
	\check m(x) = \frac{1 - c - x}{2cx} \pm \frac{i}{2cx} \sqrt{ 4cx - (1 - c - x)^2} = \frac{1 - c - x}{2cx} \pm \frac{i}{2cx} \sqrt{ (x - \lambda_-) ( \lambda_+ - x) }
\]
with the branch of $\pm$ is determined by the imaginary part of $z$ such that $\Im (z) \cdot \Im m(z) > 0$ and we recall $\lambda_- \equiv (1-\sqrt{c})^2$ and $\lambda_+ \equiv (1+\sqrt{c})^2$. For simplicity we denote
\[
	\Re \check m = \frac{1 - c - x}{2cx},\ \Im \check m = \frac1{2cx} \sqrt{ (x - \lambda_-) ( \lambda_+ - x) }
\]
and therefore
\begin{align*}
	E_{\gamma_b} &= -\frac1{2\pi i} \oint_{\gamma_b} \frac{1-f_t(z)}z \frac{ \| \bmu \|^2 m(z)}{ 1 + (\| \bmu \|^2 + c) m(z) }\ dz\\
	%%%
	&= -\frac{\| \bmu \|^2}{\pi i} \int_{\lambda_-}^{\lambda_+} \frac{1-f_t(x)}x \Im \left[\frac{ \Re \check m - i \Im \check m }{ 1 + (\| \bmu \|^2 +c) (\Re \check m - i \Im \check m )  } \right]\ dx\\
	%%%
	&= -\frac{\| \bmu \|^2}{\pi i} \int_{\lambda_-}^{\lambda_+} \frac{1-f_t(x)}x \Im \left[\frac {\Re \check m + \frac{\| \bmu \|^2 +c}{cx} - i \Im \check m}{ 1 + 2 (\| \bmu \|^2 +c) \Re \check m + \frac{(\| \bmu \|^2 +c)^2}{cx} }\right] \  dx
\end{align*}
with $z = x \pm i \varepsilon $ and $\varepsilon \to 0$ (on different sides of the real axis) and the fact that $(\Re \check m)^2 + (\Im \check m)^2 = \frac1{cx}$. We take the imaginary part and result in
\begin{equation}\label{eq:E-gamma-b}
	E_{\gamma_b} = \frac{\| \bmu \|^2}{\pi} \int_{\lambda_-}^{\lambda_+} \frac{1-f_t(x)}x \frac{  \Im \check m }{ 1 + 2 (\| \bmu \|^2 +c) \Re \check m + \frac{(\| \bmu \|^2 +c)^2}{cx} }\ dx = \frac1{2\pi} \int_{\lambda_-}^{\lambda_+} \frac{1-f_t(x)}x \frac{ \sqrt{4cx - (1-c-x)^2}}{ \lambda_s - x}\ dx 
\end{equation}
where we recall the definition $\lambda_s \equiv c+1 + \| \bmu\|^2 + \frac{c}{ \| \bmu\|^2 }$. Ultimately we assemble \eqref{eq:E-gamma-s} and \eqref{eq:E-gamma-b} to get the expression in \eqref{eq:E}. The derivations of \eqref{eq:Var}-\eqref{eq:Var-star} follow the same arguments and are thus omitted here.

% \subsection{Discussions on the Significance of Isolated Eigenpairs for Classification}
% \label{sm:intuition-eigenpair}

% compare to clustering, (re)partition of information...

\end{document}